\documentclass{article}

\PassOptionsToPackage{numbers, compress}{natbib}

\usepackage[preprint]{neurips_2023}

\usepackage[utf8]{inputenc} 
\usepackage[T1]{fontenc}    
\usepackage{hyperref}       
\usepackage{url}            
\usepackage{booktabs}       
\usepackage{amsfonts}       
\usepackage{nicefrac}       
\usepackage{microtype}      
\usepackage{xcolor}         
\usepackage{amsmath}
\usepackage{amssymb}
\usepackage{amsthm}
\usepackage{bm}
\usepackage[ruled,linesnumbered]{algorithm2e}
\usepackage{graphicx}
\usepackage{comment}
\usepackage{multirow}
\usepackage{bigstrut}
\usepackage{rotating}
\usepackage{tabularx}
\usepackage{graphicx}
\usepackage{epigraph}
\usepackage{tikz-qtree}
\usetikzlibrary{trees,decorations.text,math,calc, positioning, arrows.meta}
\usepackage{tcolorbox}
\usepackage{changepage}
\usepackage{titlesec}

\definecolor{Red}{RGB}{190,20,42}
\definecolor{CY}{RGB}{233,246,254}

\theoremstyle{definition}
\newtheorem{definition}{Definition}
\theoremstyle{theorem}

\theoremstyle{proof}

\theoremstyle{remark}
\newtheorem*{remark}{Remark}

\titlespacing*{\section}{0pt}{4pt}{4pt}

\titlespacing*{\subsection}{0pt}{3pt}{3pt}

\titlespacing*{\subsubsection}{0pt}{3pt}{3pt}

\title{
	\raisebox{-0.2cm}{\includegraphics[width=0.8cm]{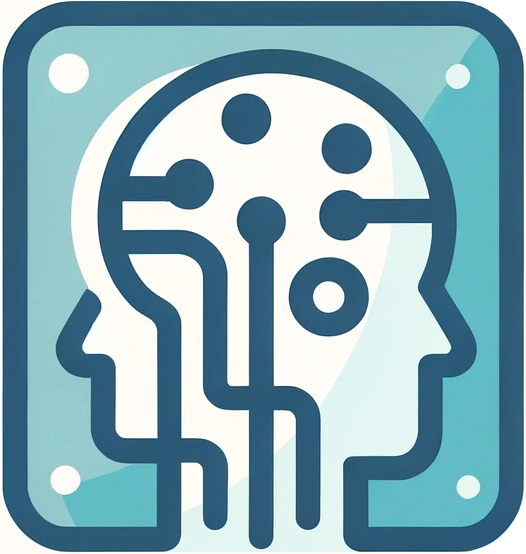}\hspace{0.1cm}}
	A Survey on the Memory Mechanism of Large Language Model based Agents
}

\author{
	Zeyu Zhang$^1$, Xiaohe Bo$^1$, Chen Ma$^1$, Rui Li$^1$,
	Xu Chen$^1$,
	Quanyu Dai$^2$,   \\
	\textbf{Jieming Zhu$^2$, Zhenhua Dong$^2$, Ji-Rong Wen$^1$}\\
	$^1$Gaoling School of Artificial Intelligence,
	Renmin University of China, Beijing, China\\
	$^2$Huawei Noah’s Ark Lab, China\\
	\texttt{zeyuzhang@ruc.edu.cn}, \texttt{xu.chen@ruc.edu.cn} \\
}

\begin{document}
	\maketitle
	\begin{abstract}
		Large language model (LLM) based agents have recently attracted much attention from the research and industry communities. Compared with original LLMs, LLM-based agents are featured in their self-evolving capability, which is the basis for solving real-world problems that need long-term and complex agent-environment interactions. 
		The key component to support agent-environment interactions is the memory of the agents. 
		While previous studies have proposed many promising memory mechanisms, they are scattered in different papers, and there lacks a systematical review to summarize and compare these works from a holistic perspective, failing to abstract common and effective designing patterns for inspiring future studies.
		To bridge this gap, in this paper, we propose a comprehensive survey on the memory mechanism of LLM-based agents.
		In specific, we first discuss ``what is'' and ``why do we need'' the memory in LLM-based agents. Then, we systematically review previous studies on how to design and evaluate the memory module. 
		In addition, we also present many agent applications, where the memory module plays an important role. 
		At last, we analyze the limitations of existing work and show important future directions.
		To keep up with the latest advances in this field, we create a repository at \url{https://github.com/nuster1128/LLM_Agent_Memory_Survey}.

	\end{abstract}
	
	\begin{figure}[h]
		\centering
		\vspace{-0.2cm}
		\includegraphics[width=0.84\textwidth]{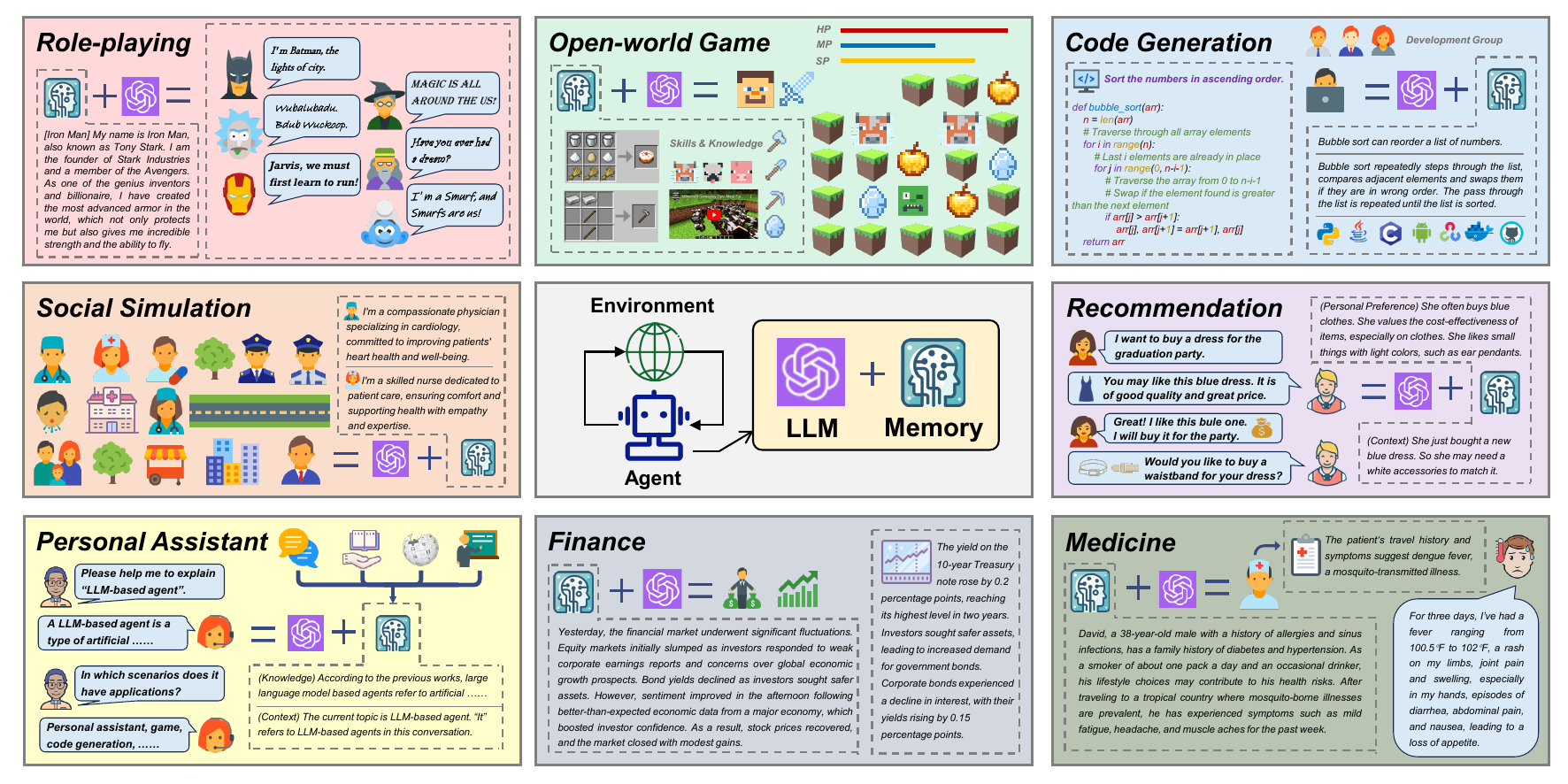}
		\vspace{-0.2cm}
		\caption{The importance of the memory module in LLM-based agents.}
		\vspace{-0.2cm}
		\label{imp}
	\end{figure}
	
	\setcounter{footnote}{0}
	
	\clearpage
	\tableofcontents
	\clearpage
	
	\section{Introduction}
	
	\setlength{\epigraphwidth}{.98\textwidth}
	
	\epigraph{\textit{"Without memory, there is no culture. Without memory, there would be no civilization, no society, no future."}}{Elie Wiesel, 1928-2016}

	Recently, large language models (LLMs) have achieved remarkable success in a large number of domains, ranging from artificial intelligence and software engineering to education and social science~\cite{chatdev,s3,survey_llm_agent1}.
	Original LLMs usually accomplish different tasks without interacting with environments.
	However, to achieve the final goal of artificial general intelligence (AGI), 
	intelligent machines should be able to improve themselves by autonomously exploring and learning from the real world. 
	{For example, if a trip-planning agent intends to book a ticket, it should send an order request to the ticket website, and observe the response before taking the next action.
		A personal assistant agent should adjust its behaviors according to the user's feedback, providing personalized responses to improve user's satisfaction.}
	To further push the boundary of LLMs towards AGI, recent years have witnessed a large number of studies on LLM-based agents~\cite{survey_llm_agent1,survey_llm_agent2}, where the key is to equip LLMs with additional modules to enhance their self-evolving capability in real-world environments.
	
	Among all the added modules, memory is a key component that differentiates the agents from original LLMs, making an agent truly an agent (see \textbf{Figure~\ref{imp}}). It plays an extremely important role in determining how the agent accumulates knowledge, processes historical experience, retrieves informative knowledge to support its actions, and so on. 
	Around the memory module, people have devoted much effort to designing its information sources, storage forms, and operation mechanisms.
	For example,~\citet{reflexion} incorporate both in-trial and cross-trial information to build the memory module for enhancing the agent's reasoning capability.
	\citet{memorybank} store memory information in the form of natural languages, which is explainable and friendly to the users.
	\citet{ret_llm} design both memory reading and writing operations to interact with environments for task solving.
	
	While previous studies have designed many promising memory modules, there still lacks a systemic study to view the memory modules from a holistic perspective.  
	To bridge this gap, in this paper, we comprehensively review previous studies to present clear taxonomies and key principles for designing and evaluating the memory module. 
	In specific, we discuss three key problems including: 
	(1) what is the memory of LLM-based agents?
	(2) why do we need the memory in LLM-based agents? 
	and (3) how to implement and evaluate the memory in LLM-based agents?
	To begin with, we detail the concepts of memory in LLM-based agents, providing both narrow and broad definitions.
	Then, we analyze the necessity of memory in LLM-based agents, showing its importance from three perspectives including cognitive psychology, self-evolution, and agent applications.
	Based on the problems of ``what'' and ``why'', we present commonly used strategies to design and evaluate the memory modules.
	For the memory design, we discuss previous works from three dimensions, that is, memory sources, memory forms, and memory operations.
	For the memory evaluation, we introduce two widely used approaches including direct evaluation and indirect evaluation via specific agent tasks.
	Next, we discuss agent applications including role-playing, social simulation, personal assistant, open-world games, code generation, recommendation, and expert systems, in order to show the importance of the memory module in practical scenarios.    
	At last, we analyze the limitations of existing work and highlight significant future directions.

	The main contributions of this paper can be summarized as follows:
	(1) We formally define the memory module and comprehensively analyze its necessity for LLM-based agents.
	(2) We systematically summarize existing studies on designing and evaluating the memory module in LLM-based agents, providing clear taxonomies and intuitive insights.
	(3) We present typical agent applications to show the importance of the memory module in different scenarios.    
	(4) We analyze the key limitations of existing memory modules and show potential solutions for inspiring future studies.
	To our knowledge, this is the first survey on the memory mechanism of LLM-based agents.
	
	The rest of this survey is organized as follows. First, we provide a systematical meta-survey for the fields of LLMs and LLM-based agents in \textbf{Section~\ref{sec:related_survey}}, categorizing different surveys and summarizing their key contributions.
	Then, we discuss the problems of ``what is'', ``why do we need'' and ``how to implement and evaluate'' the memory module in LLM-based agents in \textbf{Section~\ref{sec:definition} to~\ref{sec:evaluation}}. Next, we show the applications of memory-enhanced agents in \textbf{Section~\ref{sec:application}}.
	The discussions of the limitations of existing work and future directions come at last in \textbf{Section~\ref{sec:future}} and \textbf{Section~\ref{sec:conclusion}}.
	
	\section{Related Surveys}
	\label{sec:related_survey}

	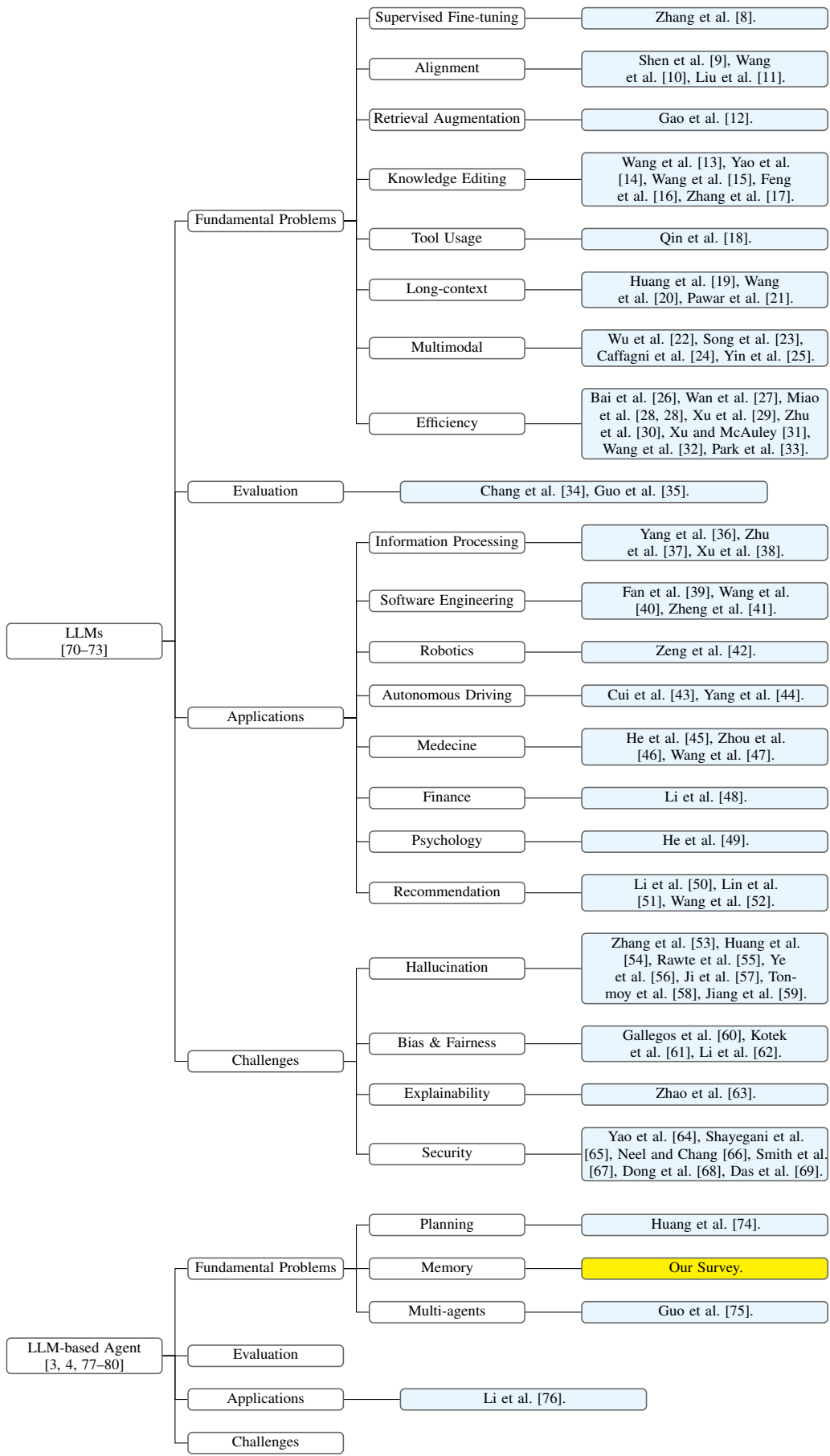
\begin{figure}[p]
		\vspace{-2cm}
		\begin{tikzpicture}
			\tikzset{
				grow'=right,level distance=30mm, sibling distance =3.5mm,
				execute at begin node=\strut,
				every tree node/.style={
					draw=gray!80!black,
					line width=0.6pt,
					text width=2.5cm,
					rounded corners=2pt,
					anchor = west,
					fill=white,
					minimum width=2mm,
					inner sep=1pt,
					align=center,
					font = {\scriptsize}},
				edge from parent/.style={draw=black,
					edge from parent fork right}
			}
			\begin{scope}[frontier/.style={sibling distance=4em,level distance = 10em}]
				\Tree
				[.{LLMs \\ \cite{survey_llm1, survey_llm2, survey_llm3,survey_llm_augment}}
				[.{Fundamental Problems}
				[.{Supervised Fine-tuning} 
				\node[fill=CY,text width=4cm](t1){ \citet{survey_sft_llm}.};
				]
				[.{Alignment}
				\node[fill=CY,text width=4cm](t1){\citet{survey_llm_alignment1, survey_llm_alignment2,survey_llm_alignment3}.};
				]
				[.{Retrieval Augmentation}
				\node[fill=CY,text width=4cm](t1){\citet{survey_retrieval_llm}.};
				]
				[.{Knowledge Editing} 
				\node[fill=CY,text width=4cm](t1){\citet{survey_knowledge_editing_llm, survey_knowledge_editing_llm2, survey_knowledge_editing_llm3, survey_knowledge_editing_llm4,survey_knowledge_editing_llm5}.};
				]
				[.{Tool Usage} 
				\node[fill=CY,text width=4cm](t1){\citet{survey_tool_learning}.};
				]
				[.{Long-context}
				\node[fill=CY,text width=4cm](t1){\citet{survey_long_context_llm, survey_long_context_llm2,survey_long_context_llm3}.};
				]
				[.{Multimodal}
				\node[fill=CY,text width=4cm](t1){\citet{survey_llm_multimodal1, survey_llm_multimodal2, survey_llm_multimodal3, survey_llm_multimodal4}.};
				]
				[.{Efficiency} 
				\node[fill=CY,text width=4cm](t1){\citet{survey_llm_efficiency1, survey_llm_efficiency2, survey_llm_efficiency3, survey_llm_efficiency3, survey_llm_efficiency5,survey_llm_efficiency6, survey_llm_efficiency7, survey_llm_efficiency8, survey_llm_efficiency9}.};
				]    				
				]
				[.{Evaluation}
				\node[fill=CY,text width=6cm](t1){\citet{survey_llm_evaluation, survey_llm_evaluation2}.};
				]
				[.{Applications}
				[.{Information Processing}
				\node[fill=CY,text width=4cm](t1){\citet{survey_llm_apply,survey_llm_apply_info1, survey_llm_apply_info2}.};
				]
				[.{Software Engineering}
				\node[fill=CY,text width=4cm](t1){\citet{survey_llm_apply_software1, survey_llm_apply_software2, survey_llm_apply_software3}.};
				]
				[.{Robotics}
				\node[fill=CY,text width=4cm](t1){\citet{survey_llm_apply_robotics}.};
				]
				[.{Autonomous Driving}
				\node[fill=CY,text width=4cm](t1){\citet{survey_llm_apply_driving1, survey_llm_apply_driving2}.};
				]
				[.{Medecine}
				\node[fill=CY,text width=4cm](t1){\citet{survey_llm_apply_med1, survey_llm_apply_med2, survey_llm_apply_med3}.};
				]
				[.{Finance}
				\node[fill=CY,text width=4cm](t1){\citet{survey_llm_apply_finance}.};
				]
				[.{Psychology}
				\node[fill=CY,text width=4cm](t1){\citet{survey_llm_apply_psy}.};
				]
				[.{Recommendation}
				\node[fill=CY,text width=4cm](t1){\citet{survey_llm_apply_recsys,survey_llm_apply_recsys2,survey_llm_apply_recsys3}.};
				]
				]
				[.{Challenges} 
				[.{Hallucination}
				\node[fill=CY,text width=4cm](t1){\citet{survey_llm_hallucination1, survey_llm_hallucination2, survey_llm_hallucination3, survey_llm_hallucination4, survey_llm_hallucination5, survey_llm_hallucination6, survey_llm_hallucination7}.};
				]
				[.{Bias \& Fairness}
				\node[fill=CY,text width=4cm](t1){\citet{survey_llm_bias_fairness1, survey_llm_bias_fairness2,survey_llm_bias_fairness3}.};
				]
				[.{Explainability}
				\node[fill=CY,text width=4cm](t1){\citet{survey_llm_explainability}.};
				]
				[.{Security}
				\node[fill=CY,text width=4cm](t1){\citet{survey_llm_security1, survey_llm_security2, survey_llm_security3, survey_llm_security4, survey_llm_security5, survey_llm_security6}.};
				]
				]
				]
			\end{scope}
		\end{tikzpicture}
		
		\vspace{0.5cm}
		\begin{tikzpicture}
			\tikzset{
				grow'=right,level distance=30mm, sibling distance =3.5mm,
				execute at begin node=\strut,
				every tree node/.style={
					draw=gray!80!black,
					line width=0.6pt,
					text width=2.5cm,
					rounded corners=2pt,
					anchor = west,
					fill=white,
					minimum width=2mm,
					inner sep=1pt,
					align=center,
					font = {\scriptsize}},
				edge from parent/.style={draw=black,
					edge from parent fork right}
			}
			\begin{scope}[frontier/.style={sibling distance=4em,level distance = 10em}]
				\Tree
				[.{LLM-based Agent \\ \cite{survey_llm_agent1,survey_llm_agent2,survey_llm_agent5,survey_llm_agent6,survey_llm_agent7,survey_llm_agent8}}
				[.{Fundamental Problems}
				[.{Planning} 
				\node[fill=CY,text width=4cm](t1){ \citet{survey_llm_agent_planning}.};
				]
				[.{Memory} 
				\node[fill=yellow,text width=4cm](t1){Our Survey.};
				]
				[.{Multi-agents} 
				\node[fill=CY,text width=4cm](t1){ \citet{survey_llm_agent4}.};
				]		
				]
				[.{Evaluation}
				]
				[.{Applications}
				\node[fill=CY,text width=4cm](t1){ \citet{survey_llm_agent3}.};
				]
				[.{Challenges} 
				]
				]
			\end{scope}
		\end{tikzpicture}
		\caption{The organization of related surveys on LLMs and LLM-based agents.}
		\vspace{-0.2cm}
		\label{fig:related_agent}
	\end{figure}
	
	In the past two years, LLMs have attracted much attention from the academic and industry communities. To systemically summarize the studies in this field, researchers have written a lot of survey papers. In this section, we briefly review these surveys (see \textbf{Figure~\ref{fig:related_agent}} for an overview), highlighting their major focuses and contributions to better position our study.

	\subsection{Surveys on Large Language Models} 
	
	In the field of LLMs,~\citet{survey_llm1} present the first comprehensive survey to summarize the background, evolution paths, model architectures, training methodologies, and evaluation strategies of LLMs. 
	~\citet{survey_llm2} and~\citet{survey_llm3} also conduct LLM surveys from the holistic view, which, however, provide different taxonomies and understandings on LLMs.
	Following these surveys, people dive into specific aspects of LLMs and review the corresponding milestone studies and key technologies. 
	These aspects can be classified into four categories including the fundamental problems, evaluation, applications, and challenges of LLMs. 
	
	\textbf{Fundamental problems}. 
	The surveys in this category aim to summarize techniques that can be leveraged to tackle fundamental problems of LLMs. 
	Specifically, \citet{survey_sft_llm} provide a comprehensive survey on the methods of supervised fine-tuning, which is a key technique for better training LLMs. 
	\citet{survey_llm_alignment1, survey_llm_alignment2} and \citet{survey_llm_alignment3} present surveys on the alignment of LLMs, which is a key requirement for LLMs to produce outputs consistent with human values. 
	\citet{survey_retrieval_llm} propose a survey on the retrieval-augmented generation (RAG) capability of LLMs, which is key to providing LLMs with factual and up-to-date knowledge and removing hallucinations. 
	\citet{survey_tool_learning} summarize the state-of-the-art methods on enabling LLMs to leverage external tools, which is fundamental for LLMs to expand their capability in domains that require specialized knowledge.
	\citet{survey_knowledge_editing_llm, survey_knowledge_editing_llm2, survey_knowledge_editing_llm3, survey_knowledge_editing_llm4} and \citet{survey_knowledge_editing_llm5} present surveys on the direction of LLM knowledge editing, which is important for customizing LLMs to satisfy specific requirements.
	\citet{survey_long_context_llm, survey_long_context_llm2} and \citet{survey_long_context_llm3} focus on long-context capabilities of LLMs, which is critical for LLMs to process more information at each time and enhance their application scenarios.  
	\citet{survey_llm_multimodal1, survey_llm_multimodal2, survey_llm_multimodal3} and \citet{survey_llm_multimodal4} summarize multi-modal LLMs, which expands the capability of LLMs from text to visual and other modalities.
	The above surveys mainly focus on the effectiveness of LLMs. Another important aspect of LLMs is their training and inference efficiency. To summarize studies on this aspect, ~\citet{survey_llm_efficiency6, survey_llm_efficiency7, survey_llm_efficiency8} and ~\citet{survey_llm_efficiency9} systematically review the techniques of model compression. 
	~\citet{survey_llm_efficiency4} and~\citet{,survey_llm_efficiency5} analyze and conclude the studies on parameter efficient fine-tuning. ~\citet{survey_llm_efficiency1,survey_llm_efficiency2,survey_llm_efficiency3} and ~\citet{survey_llm_efficiency4} {put more focuses on the efficiency of resource utilization in a general sense}. 
	
	\textbf{Evaluation}. The surveys in this category focus on how to evaluate the capability of LLMs. Specifically, \citet{survey_llm_evaluation} comprehensively summarize the evaluation methods from an overall perspective. It encompasses different evaluation tasks, methods, and benchmarks, which serve as critical parts in assessing LLM performances.
	\citet{survey_llm_evaluation2} care more about the evaluation targets and describe how to evaluate the knowledge, alignment, and safety control capabilities of LLMs, which supplement evaluation metrics beyond performance.
	
	\textbf{Applications}. The surveys in this category aim to summarize models that leverage LLMs to improve different applications. 
	More concretely, \citet{survey_llm_apply_info1} focus on the field of information retrieval (IR) and summarize studies on LLM-based query processes.
	\citet{survey_llm_apply_info2} pay more attention to information extraction (IE) and provide comprehensive taxonomies for LLM-based models in this field. 
	\citet{survey_llm_apply_recsys, survey_llm_apply_recsys2} and~\citet{survey_llm_apply_recsys3} discuss the applications of LLMs in the field of recommender system, where they utilize agents to generate data and provide recommendations. \citet{survey_llm_apply_software1,survey_llm_apply_software2}, and \citet{survey_llm_apply_software3} concentrate on how LLMs can benefit software engineering (SE) in terms of software design, development, and testing. 
	\citet{survey_llm_apply_robotics} summarize LLM-based methods in the field of robotics.
	\citet{survey_llm_apply_driving1} and~\citet{survey_llm_apply_driving2} focus on the application of autonomous driving and summarize models in this domain based on LLMs from different perspectives.
	Beyond the above domains in artificial intelligence, LLMs have also been used in natural and social science. 
	\citet{survey_llm_apply_med1, survey_llm_apply_med2} and \citet{survey_llm_apply_med3} summarize the applications of LLMs in medicine.
	\citet{survey_llm_apply_finance} focus on the applications of LLMs in finance.
	\citet{survey_llm_apply_psy} review the models on leveraging LLMs to improve the development of psychology. 
	
	\textbf{Challenges}. The surveys in this category focus on trustworthiness in LLMs, such as hallucination, bias, unfairness, explainability, security, and privacy.
	Hallucination in LLMs refers to the problem that LLMs may generate misconceptions or fabrications, impacting their reliability for downstream applications. \citet{survey_llm_hallucination1, survey_llm_hallucination2, survey_llm_hallucination3, survey_llm_hallucination4, survey_llm_hallucination5, survey_llm_hallucination6} and \citet{survey_llm_hallucination7} summarize the mainstream models for alleviating the hallucination problem in LLMs.
	The bias and unfairness problems refer to the phenomenon that LLMs may unequally treat different humans or objectives, which can lead to the propagation of societal stereotypes and discrimination. \citet{survey_llm_bias_fairness1,survey_llm_bias_fairness2} and \citet{survey_llm_bias_fairness3} comprehensively discuss these challenges and summarize existing methods for alleviating them.
	The problem of explainability means that the internal working mechanisms of LLMs are still unclear. \citet{survey_llm_explainability} systematically discuss this problem and summarize previous efforts on improving the explainability of LLMs.
	Security and privacy are also challenging problems, which have been comprehensively surveyed in \citet{survey_llm_security1, survey_llm_security2,survey_llm_security3,survey_llm_security4,survey_llm_security5} and \citet{survey_llm_security6}.
	
	\subsection{Surveys on Large Language Model-based Agents}
	Based on the capability of LLMs, people have conducted a lot of studies on building LLM-based agents, which can autonomously perceive environments, take actions, accumulate knowledge, and evolve themselves. 
	In this field,~\citet{survey_llm_agent1} present the first survey paper to systematically summarize LLM-based agents from the perspectives of agent construction, agent application, and agent evaluation. 
	~\citet{survey_llm_agent2, survey_llm_agent5,survey_llm_agent6} and~\citet {survey_llm_agent8} also summarize LLM-based agent studies from the overall perspective, but they have different focuses and taxonomies, delivering more diverse understandings on this field.
	In addition to these overall surveys, there have also emerged several papers reviewing specific aspects of LLM-based agents.
	For the fundamental problems, \citet{survey_llm_agent7} summarize studies on multi-modal agents.
	\citet{survey_llm_agent_planning} focus on the planning capability of LLM-based agents.
	\citet{survey_llm_agent4} pay more attention to the scenarios of multi-agent interactions.
	For the applications, \citet{survey_llm_agent3} provide a summarization on LLM-based agents that are leveraged as personal assistants. 
	
	\textbf{Position of this work}. Our survey summarizes the studies on a fundamental problem of LLM-based agents, that is, the agent's memory mechanism. To our knowledge, this is the first survey in this direction.
	We hope it can not only inspire more advanced memory architectures in the future, but also provide newcomers with comprehensive starting materials.    
	
	\section{What is the Memory of LLM-based Agent}
	\label{sec:definition}
	Interacting and learning from environments is a basic requirement of LLM-based agents. 
	In the agent-environment interaction process, there are three key phases, that is, (1) the agent perceives information from the environment, and stores it into the memory; (2) the agent processes the stored information to make it more usable; and (3) the agent takes the next action based on the processed memory information. 
	In all these phases, memory plays an extremely important role.
	In the following, we first define the memory of the agent from both narrow and broad perspectives, and then, detail the execution processes of the above three phases based on the memory module.
	
	\subsection{Basic Knowledge}\label{basic}
	For clear presentations, we first introduce several important background knowledge as follows: 
	
	\begin{definition}[Task]
		\label{task}
		Task is the final target that the agent needs to achieve, for example, booking a flight ticket for Alice, recommending a restaurant for Bob, and so on. Formally, we use $\mathcal{T}$ to represent a task and label different tasks by subscripts in the following contents.
	\end{definition}
	
	\begin{definition}[Environment]
		In a narrow sense, environment is the object that the agent needs to interact with to accomplish the task. 
		For the examples in definition~\ref{task}, the environments are Alice and Bob, who provide feedback on the agent's actions.
		More broadly, environment can be any contextual factors that influence the agent's decisions, such as the weather when booking flight tickets, the time and location when recommending restaurants, etc.
	\end{definition}
	
	\begin{definition}[Trial and Step]
		To accomplish a task, the agent needs to interact with the environment. Usually, the agent first takes an action, and then the environment responds to this action. At last, the agent takes the next action based on the response. This process iterates until the task is finished. The complete agent-environment interaction process is called a \textbf{{trial}}, and each interaction turn is called a \textbf{{step}}.
		For each trial, the agent can take multiple steps to form a potential solution to the task.
		For each task, the agent can explore multiple trials to accomplish the task~\cite {reflexion}.
		Formally, at step $t$, we use $a_t$ and $o_t$ to represent the agent action and the observed environment response, respectively. Then, a $T$-length trial can be represented as $\xi_T = \{a_1,o_1,a_2,o_2, ..., o_T, a_T\}$.
	\end{definition}
	In the above definitions, task and environment are the most coarse-grained concepts, while step is the most fine-grained one. They together describe the complete agent-environment interaction process.
	
	\begin{figure*}[tb]
		\centering
		\setlength{\fboxrule}{0.pt}
		\setlength{\fboxsep}{0.pt}
		\fbox{
			\includegraphics[width=\linewidth]{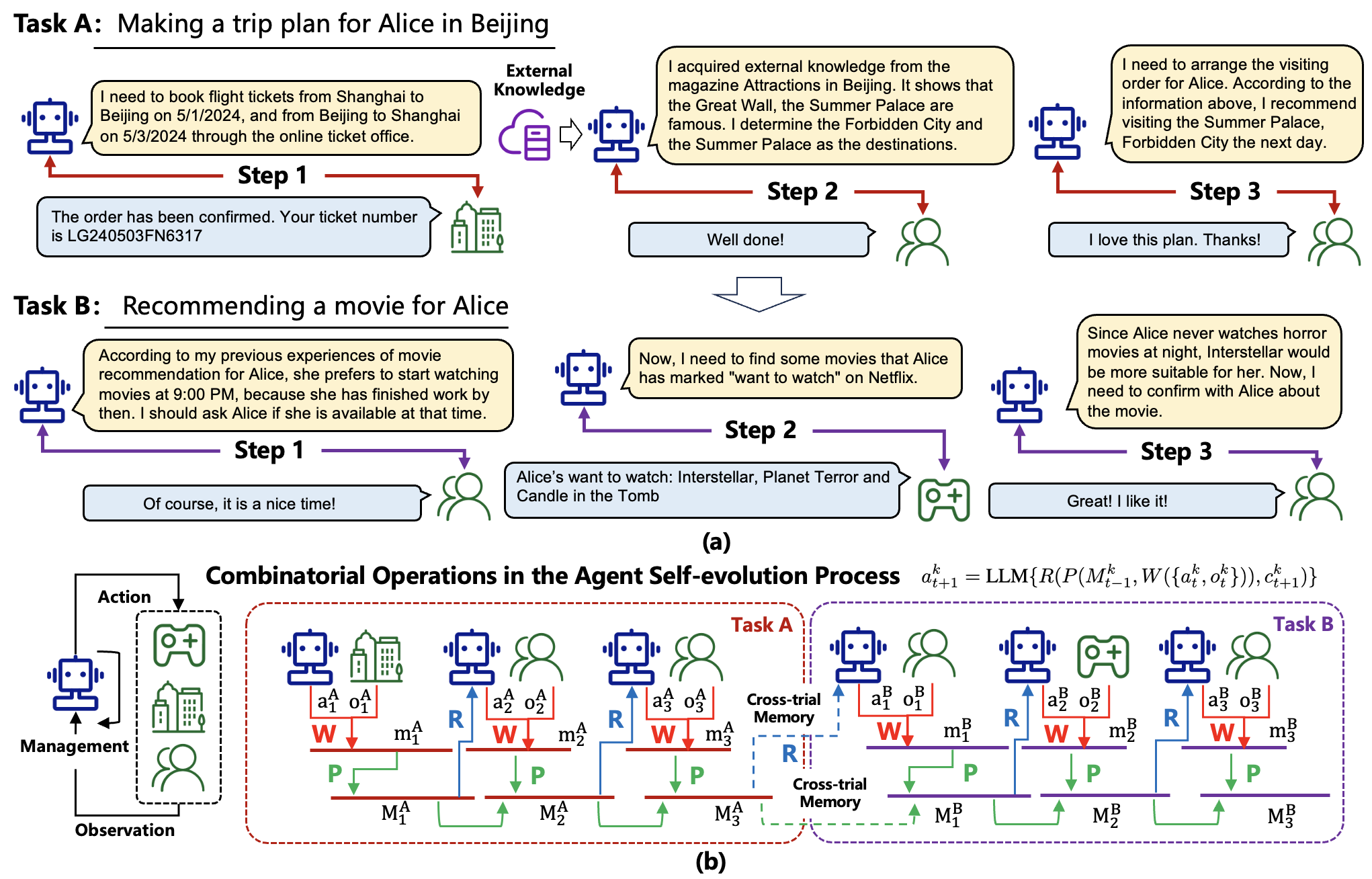}
		}
		\caption{(a) Examples of the potential trials in the agent-environment interaction process.
			(b) Illustration of the memory reading, writing, and management processes, where dotted lines mean that the cross-trial information can be incorporated into the memory module.}
		\label{fig:what}
		\vspace{-0.2cm}
	\end{figure*}
	
	\textbf{A Toy Example}.
	Based on the above definitions, we show a toy example and use it throughout the paper to provide more intuitions. 
	In this example, the agent has two tasks: 
	(A) \textbf{Making a trip plan for Alice in Beijing from 5/1/2024 to 5/3/2024}, where the agent has to (\romannumeral1) buy a round-trip flight ticket, (\romannumeral2) determine several attractions according to Alice's preferences, and (\romannumeral3) arrange the visiting order of these attractions.  
	(B) \textbf{Recommending a movie for Alice on 5/10/2024}, where the agent has to (\romannumeral1) determine the time that Alice is free to watch movies, and (\romannumeral2) select a movie according to Alice's watching preference. 
	For each task, we present a potential trial in the following (also illustrated in \textbf{Figure~\ref{fig:what}}(a)).
	For task (A), the trial is:
	
	[Step 1] \textbf{Agent}: I need to 
	book flight tickets from Shanghai to Beijing on 5/1/2024, and from Beijing to Shanghai on 5/3/2024 through the online ticket office. 
	\textbf{Environment (Online Ticket Office)}: The order has been confirmed. The departure flight takes off at 9:00 AM, and your ticket number is EK240501SX9021. The returning flight takes off at 4:30 PM, and your ticket number is LG240503FN6317. Please arrive at the airport on time for your flight.

	[Step 2] \textbf{Agent}: I have acquired external knowledge from the magazine \textit{Attractions in Beijing}. It shows that \textit{the Great Wall, the Forbidden City} and \textit{the Summer Palace} are famous. According to Alice's preferences, I think she prefers ancient architecture, so I determine \textit{the Forbidden City} and \textit{the Summer Palace} as the destinations. 
	\textbf{Environment (Alice)}: Well done!
	
	[Step 3] \textbf{Agent}: Now, I need to arrange the visiting order for Alice. According to the information above, I recommend visiting \textit{the Summer Palace} after she arrives in Beijing because it is suitable to spend a whole afternoon there and watch the sunset. Then, she can go to \textit{the Forbidden City} the next day, which would give her plenty of time. Finally, she can return to Shanghai on the third day and take a rest.
	\textbf{Environment (Alice)}: I love this plan. Thanks!
	
	For task (B), the trial is:
	
	[Step 1] \textbf{Agent}: According to my previous experiences of movie recommendations for Alice, she prefers to start watching movies at 9:00 PM, because she has finished work by then. I should ask Alice if she is available at that time.  \textbf{Environment (Alice)}: Of course, it is a nice time!
	
	[Step 2] \textbf{Agent}: Now, I need to find some movies that Alice has marked "want to watch" on Netflix. \textbf{Environment (Netflix)}: Alice's want to watch: \textit{Interstellar}, \textit{Planet Terror} and \textit{Candle in the Tomb}.
	
	[Step 3] \textbf{Agent}: Since Alice never watches horror movies at night, \textit{Interstellar} would be more suitable for her. Now, I need to confirm with Alice about the movie. \textbf{Environment (Alice)}: Great! I like it!
		
	\subsection{Narrow Definition of the Agent Memory}
	In a narrow sense, the memory of the agent is only relevant to the historical information within the same trial.
	Formally, for a given task, the historical information of the trial before step $t$ is $\xi_t = \{a_1,o_1,a_2,o_2, ..., a_{t-1},o_{t-1}\}$, and then the memory is derived based on $\xi_t$.
	{In the above toy example, for task (A), the agent at [step 3] needs to arrange the visiting order for Alice; at this time, its memory contains the information about the selected attractions and arrival time in [step 1] and [step 2]. For task (B), the agent has to choose a movie for Alice at [step 3]; at this time, its memory contains the arranged time to watch films.}
	
	\subsection{Broad Definition of the Agent Memory}
	In a broad sense, the memory of the agent can come from much wider sources, for example, the information across different trials and the external knowledge beyond the agent-environment interactions. 
	Formally, given a series of sequential tasks $\{\mathcal{T}_1,~\mathcal{T}_2,...,~\mathcal{T}_K\}$, for task $\mathcal{T}_k$, the memory information at step $t$ comes from three sources: (1) the historical information within the same trial, that is, $\xi_t^k = \{a_1^k,o_1^k, ..., a_{t-1}^k,o_{t-1}^k\}$, where we add superscript $k$ to label the task index. (2) The historical information across different trials, that is, $\Xi^{k} = \{\xi^1, \xi^2, ..., \xi^{k-1},\xi^{k'}\}$, where $\xi^j$~$(j\in\{1, ..., k-1\})$ represents the trials of task $j$\footnote{For each task, there can be multiple trials for exploring the final solution, and all of them can be incorporated into the memory.}, and $\xi^{k'}$ denotes the previously explored trials for task $\mathcal{T}_k$. (3) External knowledge, which is represented by $D_t^k$. 
	The memory of the agent is derived based on $(\xi_t^k,~\Xi^{k},~D_t^k)$.
	In the above toy example, for task (A), if there are several failed trials, that is, the feedback from Alice is negative, then these trials can be incorporated into the agent's memory to avoid future similar errors (corresponding to $\xi^{k'}$). 
	In addition, for task (B), the agent may recommend movies relevant to the attractions that Alice has visited in task (A) to capture her recent preferences (corresponding to $\{\xi^1, \xi^2, ..., \xi^{k-1}\}$).
	In the agent decision process, it has also referred to the magazine \textit{Attractions in Beijing} for making trip plans, which is the external knowledge (corresponding to $D_t^k$) for the current task $\mathcal{T}_k$.

	\subsection{Memory-assisted Agent-Environment Interaction}
	As mentioned at the beginning of \textbf{Section~\ref{sec:definition}}, there are three key phases in the agent-environment interaction process. 
	The agent memory module implements these phases through three operations including memory writing, memory management, and memory reading. 
	
	\textbf{Memory Writing}. 
	This operation aims to project the raw observations into the actually stored memory contents, which are more informative~\cite{ret_llm} and concise~\cite{memorybank}.
	It corresponds to the first phase of the agent-environment interaction process.
	Given a task $\mathcal{T}_k$, if the agent takes an action $a^k_t$ at step $t$, and the environment provides an observation $o^k_t$, then the memory writing operation can be formally represented as:
	$$m_{t}^k = W(\{a^k_t, o^k_t\}),$$
	where $W$ is a projecting function. 
	$m_t^k$ is the finally stored memory contents, which can be either natural languages or parametric representations.    
	In the above toy example, for task (A), the agent is supposed to remember the flight arrangement and the decision of attractions after [step 2].
	For task (B), the agent should memorize the fact that Alice hopes to watch movies at 9:00 PM, after [step 1].
	
	\textbf{Memory Management}. 
	This operation aims to process the stored memory information to make it more effective, for example, summarizing high-level concepts to make the agent more generalizable~\cite{memorybank}, merging similar information to reduce redundancy~\cite{ret_llm}, and forgetting unimportant or irrelevant information to remove its negative influence.
	This operation corresponds to the second phase of the agent-environment interaction process.
	Let $M_{t-1}^k$ be the memory contents for task $k$ before step $t$, and suppose $m_{t}^k$ is the stored information at step $t$ based on the above memory writing operation, then, the memory management operation can be represented by:
	$$M_{t}^k = P(M_{t-1}^k, m_{t}^k),$$
	where $P$ is a function that iteratively processes the stored memory information.
	For the narrow memory definition, the iteration only happens within the same trial, and the memory is emptied when the trial is ended. For the broad memory definition, the iteration happens across different trials or even tasks, as well as the integrations of external knowledge.
	{For task (B) in the above toy example, the agent can conclude that Alice enjoys watching science fiction movies in the evening, which can be used as a default rule to make recommendations for Alice in the future.}
	
	\textbf{Memory Reading}. 
	This operation aims to obtain important information from the memory to support the next agent action.
	It corresponds to the third phase of the agent-environment interaction process.
	Suppose $M_{t}^k$ is the memory contents for task $k$ at step $t$, $c_{t}^k$ is the context of the next action, then the memory reading operation can be represented by:
	$$\hat{M}_{t}^k = R(M_{t}^k, c_{t+1}^k),$$
	where $R$ is usually implemented by computing the similarity between $M_{t}^k$ and $c_{t+1}^k$~\cite{expel}. $\hat{M}_{t}^k$ is used as parts of the final prompt to drive the agent's next action.
	{For task (B) in the above toy example, when the agent decides on the final recommended movie in [Step 3], it should focus on the ``want to watch'' list in [Step 2] and select one from it.}
	
	Based on the above operations, we can derive a unified function for the evolving process from $\{a^k_t, o^k_t\}$ to $a^k_{t+1}$, that is:
	$$a^k_{t+1} = \text{LLM}\{R(P(M_{t-1}^k, W(\{a^k_t, o^k_t\})), c_{t+1}^k)\},$$
	where $\text{LLM}$ is the large language model.
	The complete agent-environment interaction process can be easily obtained by iteratively expanding this function (see \textbf{Figure~\ref{fig:what}}(b) for an intuitive illustration). 
	
	\begin{remark}
		This function provides a general formulation of the agent memorizing process.
		Previous works may use different specifications. 
		For example, in~\cite {reflexion}, $R$ and $P$ are set as identical functions, and $P$ only takes effect at the end of a trial.
		In \citet{generative_agents}, $R$ is implemented based on three criteria including similarity, time interval, and importance, and $P$ is realized by a reflection process to obtain more abstract thoughts.
		In this section, we focus on the overall framework of the agent's memory operations. More detailed realizations of $W$, $P$, and $R$ are deferred in \textbf{Section~\ref{sec:implement}}.
	\end{remark}

	\section{Why We Need the Memory in LLM-based Agent}
	\label{sec:necessity}
	Above, we have introduced what is the memory of LLM-based agents.
	Before comprehensively presenting how to implement it, in this section, we briefly show why memory is necessary for building LLM-based agents, where we expand our discussion from three perspectives including cognitive psychology, self-evolution, and agent applications.
	
	\subsection{Perspective of Cognitive Psychology}
	Cognitive psychology is the scientific study of human mental processes such as attention, language use, memory, perception, problem-solving, creativity, and reasoning\footnote{https://en.wikipedia.org/wiki/Cognitive\_psychology}.
	Among these processes, memory is widely recognized as an extremely important one~\cite{solso1979cognitive}. 
	It is fundamental for humans to learn knowledge by accumulating important information and abstracting high-level concepts~\cite{craik1972levels}, form social norms by remembering cultural values and individual experiences~\cite{leydesdorff2017memory}, take reasonable behaviors by imagining the potential positive and negative consequences~\cite{johnson1983mental}, and among others. 
	
	A major goal of LLM-based agents is to replace humans for accomplishing different tasks.  
	To make agents behave like humans, following human's working mechanisms to design the agents is a natural and essential choice~\cite{laird2019soar}.
	Since memory is important for humans, designing memory modules is also significant for the agents. 
	In addition, cognitive psychology has been studied for a long time, so many effective human memory theories and architectures have been accumulated, which can support more advanced capabilities of the agents~\cite{sun2001duality}. 
	
	\subsection{Perspective of Self-Evolution}
	To accomplish different practical tasks, agents have to self-evolve in dynamic environments~\cite{survey_rl}. In the agent-environment interaction process, the memory is key to the following aspects:
	(1) \textbf{Experience accumulation}.
	An important function of the memory is to remember past error plannings, inappropriate behaviors, or failed experiences, so as to make the agent more effective for handling similar tasks in the future~\cite{synapse}. 
	This is extremely important for enhancing the learning efficiency of the agent in the self-evolving process.      
	(2) \textbf{Environment exploration}. 
	To autonomously evolve in the environment, the agents have to explore different actions and learn from the feedback~\cite{montazeralghaem2020reinforcement}. By remembering historical information, the memory can help to better decide when and how to make explorations, for example, focusing more on previously failed trials or actions with lower exploring frequencies~\cite{gitm}.
	(3) \textbf{Knowledge abstraction}.
	Another important function of the memory is to summarize and abstract high-level information from raw observations, which is the basis for the agent to be more adaptive and generalizable to unseen environments~\cite{expel}.
	In summary, self-evolution is the basic characteristic of LLM-based agents, and memory is of key importance to self-evolution.
	
	\subsection{Perspective of Agent Applications}
	In many applications, memory is an indispensable component of the agent.
	For example, in a conversational agent, the memory stores information about historical conversations, which is necessary for the agent to generate the next response. Without memory, the agent does not know the context, and cannot continue the conversation~\cite{memochat}.
	In a simulation agent, memory is of great importance to make the agent consistently follow the role profiles. 
	Without memory, the agent may easily step out of the role during the simulation process~\cite{recagent}. 
	Both of the above examples show that the memory is not an optional component, but is necessary for the agents to accomplish given tasks. 
	
	In the above three perspectives, the first one reveals that the memory builds the cognitive basis of the agent. The second and third ones show that the memory is necessary for the agent's evolving principles and applications, which provide insights for designing agents with memory mechanisms. 
	
	\section{How to Implement the Memory of LLM-based Agent}
	\label{sec:implement}
\begin{figure*}[tb]
		\centering
		\setlength{\fboxrule}{0.pt}
		\setlength{\fboxsep}{0.pt}
		\fbox{
			\includegraphics[width=\linewidth]{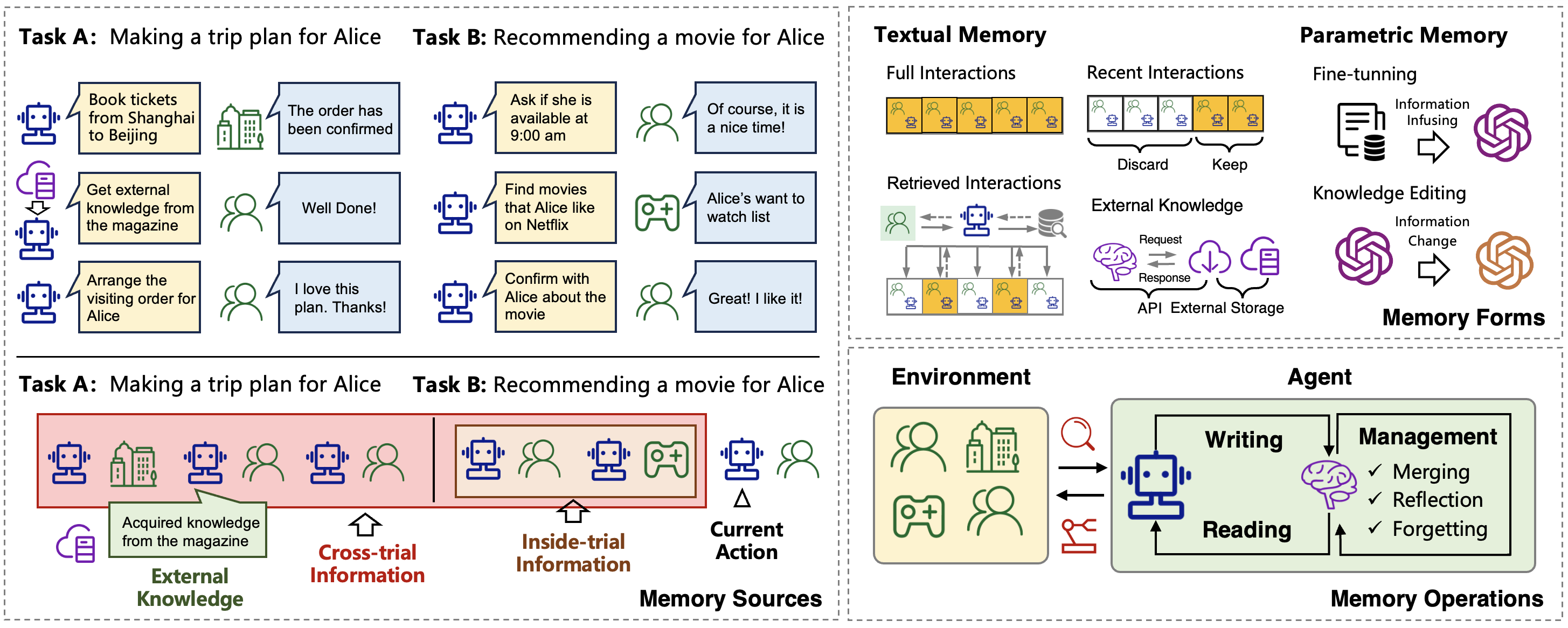}
		}
		\caption{An overview of the sources, forms, and operations of the memory in LLM-based agents.}
		\label{fig:how}
		\vspace{-0.6cm}
	\end{figure*}

	In this section, we discuss the implementation of the memory module from three perspectives: memory sources, memory forms, and memory operations. 
	Memory sources refer to where the memory contents come from. 
	Memory forms focus on how to represent the memory contents.
	Memory operations aim to process the memory contents. These three perspectives provide a comprehensive review of memory implementation methods, which is helpful for future research. For better demonstration, we present an overview of implementation methods in \textbf{Figure~\ref{fig:how}}.
	
	\subsection{Memory Sources}
	In previous works, the memory contents may come from different sources. Based on our formulation in \textbf{Section~\ref{sec:definition}}, these sources can be classified into three categories, that is, the information inside a trial, the information across different trials, and the external knowledge. The former two are dynamically generated in the agent-environment interaction process (\emph{e.g.}, task internal information), while the latter is static information outside the loop (\emph{e.g.}, task external information). We summarize previous works on memory sources in \textbf{{Table~\ref{tab:memory_source}}}.
	
	\renewcommand{\arraystretch}{1.34}
	\begin{table}[tbp]
		\centering
		\caption{Summarization of the memory sources. We use \checkmark~and \texttimes~to label whether or not the corresponding source is adopted in the model.}
		\resizebox{\linewidth}{!}{
			\begin{tabular}{cccc}
				\hline
				\hline
				\textbf{Models} & \multicolumn{1}{c}{\textbf{Inside-trial Information}} & \multicolumn{1}{c}{\textbf{Cross-trial Information}} & \multicolumn{1}{c}{\textbf{External Knowledge}} \bigstrut\\
				\hline
				MemoryBank~\cite{memorybank} & \checkmark & \texttimes & \texttimes \bigstrut[t]\\
				RET-LLM~\cite{ret_llm} & \checkmark & \texttimes & \checkmark \\
				ChatDB~\cite{chatdb} & \checkmark & \texttimes & \checkmark \\
				TiM~\cite{tim} & \checkmark & \texttimes & \texttimes \\
				SCM~\cite{scm} & \checkmark & \texttimes & \texttimes \\
				Voyager~\cite{voyager} & \checkmark & \texttimes & \texttimes \\
				MemGPT~\cite{memgpt} & \checkmark & \texttimes & \texttimes \\
				MemoChat~\cite{memochat} & \checkmark & \texttimes & \texttimes \\
				MPC~\cite{mpc} & \checkmark & \texttimes & \texttimes \\
				Generative Agents~\cite{generative_agents} & \checkmark & \texttimes & \texttimes \\
				RecMind~\cite{recmind} & \checkmark & \texttimes & \checkmark \\
				Retroformer~\cite{retroformer} & \checkmark & \checkmark & \checkmark \\
				ExpeL~\cite{expel} & \checkmark & \checkmark & \checkmark \\
				Synapse~\cite{synapse} & \checkmark & \checkmark & \texttimes \\
				GITM~\cite{gitm} & \checkmark & \checkmark & \checkmark \\
				ReAct~\cite{react} & \checkmark & \texttimes & \checkmark \\
				Reflexion~\cite{reflexion} & \checkmark & \checkmark & \checkmark \\
				RecAgent~\cite{recagent} & \checkmark & \texttimes & \texttimes \\
				Character-LLM~\cite{character_llm} & \checkmark & \texttimes & \checkmark \\
				MAC~\cite{mac} & \checkmark & \texttimes & \texttimes \\
				Huatuo~\cite{huatuo} & \checkmark & \texttimes & \checkmark \\
				ChatDev~\cite{chatdev} & \checkmark & \texttimes & \texttimes \\
				InteRecAgent~\cite{interecagent} & \checkmark & \texttimes & \checkmark \\
				MetaAgents~\cite{metaagents} & \checkmark & \texttimes & \texttimes \\
				TPTU~\cite{tptu, tptu_v2} & \checkmark & \texttimes & \checkmark \\
				MetaGPT~\cite{metagpt} & \checkmark & \checkmark & \texttimes \\
				S$^3$~\cite{s3} & \checkmark & \texttimes & \texttimes \\
				InvestLM~\cite{investlm} & \checkmark & \texttimes & \checkmark \bigstrut[b]\\
				
				\hline
				\hline
			\end{tabular}%
		}
		\label{tab:memory_source}
		\vspace{-0.6cm}
	\end{table}
	
	\subsubsection{Inside-trial Information}
	In the agent-environment interaction process, the historical steps within a trial are usually the most relevant and informative signals to support the agent's future actions.
	Almost all the previous works use this information as a part of the memory sources.
	
	\textbf{Representative Studies}.
	Generative Agents~\cite{generative_agents} aims to simulate human's daily behaviors by using LLM-based agents. The memory of an agent is derived from the historical behaviors to achieve a target, for example, the collection of relevant papers when researching on a specific topic.
	MemoChat~\cite{memochat} aims to chat with humans, where the memory of the agent is derived based on the conversation history of a dialogue session.
	TiM~\cite{tim} aims to enhance the agent's reasoning capability by self-generating multiple thoughts after accomplishing a task, which is used as the memory to provide more generalizable information.
	Voyager~\cite{voyager} focuses on building game agents based on Minecraft, where the memory contains executable codes of preliminary and basic actions to accomplish a task.
	It should be noted that the inside-trial information not only includes agent-environment interactions, but also contains interaction contexts, such as time and location information.
	
	\textbf{Discussion}.
	The inside-trial information is the most obvious and intuitive source that should be leveraged to construct the agent's memory since it is highly relevant to the current task that the agent has to accomplish. However, relying solely on inside-trial information may prevent the agent from accumulating valuable knowledge from various tasks and learning more generalizable information.
	Thus, many studies also explore how to effectively utilize the information across different tasks to build the memory module, which is detailed in the following sections.
	
	\subsubsection{Cross-trial Information}
	For LLM-based agents, the information accumulated across multiple trials in the environment is also a crucial part of the memory, typically including successful and failed actions and their insights, such as failure reasons, common action patterns to succeed, and so on. 
	
	\textbf{Representative Studies}.
	One of the most prominent studies is Reflexion~\cite{reflexion}, which proposes verbal reinforcement learning for LLM-based agents. It derives the experiences from past trials in verbal form, and applies them in subsequent trials to improve the performance of the same task.
	Furthermore, Retroformer~\cite{retroformer} fine-tunes the reflection model, enabling the agent to extract cross-trial information from past trials more effectively.
	In Synapse~\cite{synapse}, the agents focus on solving the computer control tasks. Their memory can record cross-trial information through successful exemplars, which would be used as references on similar trials. In ExpeL~\cite{expel}, the agents are required to solve a collection of complex interactive tasks within the environment.
	They store and organize completed trajectories, and recall similar ones for the new task. In the recalled trajectories, successful cases will be compared with failed ones to identify the patterns to succeed.
	
	\textbf{Discussion}.
	According to the accumulated memory of cross-trial information, the agents are able to accumulate experiences, which is important for their evolution.
	Based on the past experiences, the agents can adjust their actions based on the overall feedback of the whole process. In contrast to the inside-trial observations, which serve as short-term memory, the trial experiences can be considered as long-term memory. It utilizes feedback from different trials to support a wider range of agent trials, providing more prolonged experiential support for agents.
	However, the limitation lies in the fact that both inside-trial and cross-trial information require the agents to personally engage in agent-environment interactions, where external experiences and knowledge are not included.
	
	\subsubsection{External Knowledge}
	An important characteristic of LLM-based agents is that they can be directly communicated and controlled in natural languages. 
	As such, LLM-based agents can easily incorporate external knowledge in textual forms (\emph{e.g.}, Wikipedia~\footnote{\url{https://www.wikipedia.org}}) to facilitate their decisions.
	
	\textbf{Representative Studies}.
	In ReAct~\cite{react}, the agents are required to answer questions about general knowledge by multiple reasoning steps. They can utilize Wikipedia APIs to obtain external knowledge if they lack information during these steps. 
	GITM~\cite{gitm} intends to design agents in Minecraft, which can explore in complex and sparse-reward environments. The agents draw from the online Minecraft Wiki and craft recipes to provide an infinite source of knowledge for their navigation.
	CodeAgent~\cite{codeagent} focuses on the repo-level code generation task, which commonly requires complex dependencies and extensive documentation. It designs a web search strategy for acquiring related external knowledge. ChatDoctor~\cite{chatdoctor} adapts LLM-based agents to the medical domain. It fine-tunes an acquisition process to retrieve external knowledge from Wikipedia and medical databases.
	
	\textbf{Discussion}.
	The external knowledge can be obtained from both private and public sources. It provides LLM-based agents with much knowledge beyond their internal environment, which might be difficult or even impossible for the agent to acquire by agent-environment interactions. Moreover, most external knowledge can be acquired by accessing the APIs of various tools dynamically in real time according to the task needs, thus mitigating the problem of outdated knowledge.
	Integrating external knowledge into the memory of LLM-based agents significantly expands their knowledge boundaries, providing them with unlimited, up-to-date, and well-founded knowledge for decision-making.

	\subsection{Memory Forms}
	In general, there are two forms to represent the memory contents: textual form and parametric form. 
	In textual form, the information is explicitly retained and recalled by natural languages. 
	In parametric form, the memory information is encoded into parameters and implicitly influences the agent's actions.
	We summarize previous works on memory forms with their implementations in \textbf{Table~\ref{tab:memory_form}}.
	
	\begin{table}[tbp]
		\centering
		\vspace{-0.1cm}
		\caption{Summarization of the memory forms. We use \checkmark~and \texttimes~to label whether or not the corresponding memory form is adopted in the model.
		}
		\resizebox{0.99\linewidth}{!}{
			\begin{tabular}{ccccccc}
				\hline
				\hline
				\multirow{2}[4]{*}{\textbf{Models}} & \multicolumn{4}{c}{\textbf{Textual Form}} & \multicolumn{2}{c}{\textbf{Parametric Form}} \bigstrut\\
				\cline{2-7}          & \textbf{\ Complete\ } & \textbf{Recent} & \textbf{Retrieved} & \textbf{External} & \textbf{Fine-tuning} & \textbf{Editing} \bigstrut\\
				\hline
				MemoryBank~\cite{memorybank} & \texttimes & \texttimes & \checkmark & \texttimes & \texttimes & \texttimes \bigstrut[t]\\
				RET-LLM~\cite{ret_llm} & \texttimes & \texttimes & \checkmark & \texttimes & \texttimes & \texttimes \\
				ChatDB~\cite{chatdb} & \texttimes & \texttimes & \checkmark & \texttimes & \texttimes & \texttimes \\
				TiM~\cite{tim} & \texttimes & \texttimes & \checkmark & \texttimes & \texttimes & \texttimes \\
				SCM~\cite{scm} & \texttimes & \checkmark & \checkmark & \texttimes & \texttimes & \texttimes \\
				Voyager~\cite{voyager} & \texttimes & \texttimes & \checkmark & \texttimes & \texttimes & \texttimes \\
				MemGPT~\cite{memgpt} & \texttimes & \checkmark & \checkmark & \texttimes & \texttimes & \texttimes \\
				MemoChat~\cite{memochat} & \texttimes & \texttimes & \checkmark & \texttimes & \texttimes & \texttimes \\
				MPC~\cite{mpc} & \texttimes & \texttimes & \checkmark & \texttimes & \texttimes & \texttimes \\
				Generative Agents~\cite{generative_agents} & \texttimes & \texttimes & \checkmark & \texttimes & \texttimes & \texttimes \\
				RecMind~\cite{recmind} & \checkmark & \texttimes & \texttimes & \texttimes & \texttimes & \texttimes \\
				Retroformer~\cite{retroformer} & \checkmark & \texttimes & \texttimes & \checkmark & \checkmark & \texttimes \\
				ExpeL~\cite{expel} & \checkmark & \texttimes & \checkmark & \checkmark & \texttimes & \texttimes \\
				Synapse~\cite{synapse} & \texttimes & \texttimes & \checkmark & \texttimes & \texttimes & \texttimes \\
				GITM~\cite{gitm} & \checkmark & \texttimes & \checkmark & \checkmark & \texttimes & \texttimes \\
				ReAct~\cite{react} & \checkmark & \texttimes & \texttimes & \checkmark & \texttimes & \texttimes \\
				Reflexion~\cite{reflexion} & \checkmark & \texttimes & \texttimes & \checkmark & \texttimes & \texttimes \\
				RecAgent~\cite{recagent} & \texttimes & \checkmark & \checkmark & \texttimes & \texttimes & \texttimes \\
				Character-LLM~\cite{character_llm} & \texttimes & \checkmark & \texttimes & \texttimes & \checkmark & \texttimes \\
				MAC~\cite{mac} & \texttimes & \texttimes & \texttimes & \texttimes & \texttimes & \checkmark \\
				Huatuo~\cite{huatuo} & \checkmark & \texttimes & \texttimes & \texttimes & \checkmark & \texttimes \\
				ChatDev~\cite{chatdev} & \checkmark & \texttimes & \texttimes & \texttimes & \texttimes & \texttimes \\
				InteRecAgent~\cite{interecagent} & \texttimes & \checkmark & \checkmark & \checkmark & \texttimes & \texttimes \\
				MetaAgents~\cite{metaagents} & \texttimes & \texttimes & \checkmark & \texttimes & \texttimes & \texttimes \\
				TPTU~\cite{tptu, tptu_v2} & \checkmark & \texttimes & \texttimes & \checkmark & \texttimes & \texttimes \\
				MetaGPT~\cite{metagpt} & \checkmark & \texttimes & \texttimes & \texttimes & \texttimes & \texttimes \\
				S$^3$~\cite{s3} & \texttimes & \texttimes & \checkmark & \texttimes & \texttimes & \texttimes \\
				InvestLM~\cite{investlm} & \checkmark & \texttimes & \texttimes & \texttimes & \checkmark & \texttimes \bigstrut[b]\\
				\hline
				\hline
			\end{tabular}%
		}
		\label{tab:memory_form}%
		\vspace{-0.8cm}
	\end{table}%
	
	\subsubsection{Memory in Textual Form}
	Textual form is currently the mainstream method to represent the memory contents, which is featured in better interpretability, easier implementation, and faster read-write efficiency.
	In specific, the textual form can be both non-structured representations like raw natural languages and structured information such as tuples, databases, and so on. 
	In general, previous studies use the textual form memory to store four types of information including (1) complete agent-environment interactions, (2) recent agent-environment interactions, (3) retrieved agent-environment interactions, and (4) external knowledge.
	In the former three methods, the memory leverages natural languages to describe the information within the agent-environment interaction loop.
	In the former three types, they record the information inside the agent-environment interaction loop, while the last type leverages natural languages to store information outside that loop.
	
	\textbf{Complete Interactions}.
	This method stores all the information of the agent-environment interaction history based on long-context strategies~\cite{longchat}.
	For the example in \textbf{Section~\ref{basic}},  the memory of the agent in task (A) after step 2 can be implemented by concatenating all the information before step 2, and the final textual form memory is: "Your memory is [Step 1] (Agent) ... (Online Ticket Office) ... [Step 2] ... Please infer based on your memory". 
	
	In the previous work, different models store the memory information using different strategies.
	For example, in LongChat~\cite{longchat}, the agents focus on understanding natural languages in long-context scenarios. It fine-tunes the foundation model for better adapting to memorize complete interactions.
	Memory Sandbox~\cite{memory_sandbox} intends to alleviate the impact of irrelevant memory in conversations. It designs a transparent and interactive method to manage the memory of agents, which removes irrelevant memory before concatenating them as a prompt.
	Moreover, some efforts are dedicated to enhancing the capacity of LLMs to handle longer contexts~\cite{giraffe,longllama}.
	
	While storing all the agent-environment interactions can maintain comprehensive information, obvious limitations exist in terms of computational cost, inference time, and inference robustness.
	Firstly, the fast-growing long-context memory in practice results in high computational cost during LLM inference, due to the quadratic growth of the time complexity of attention computation with sequence length.
	It thus requires much more computing resources and significantly increases inference latency, which hinders its practical deployment.
	What's more, with its fast growth, the memory length can easily exceed the upper bound of the sequence length during LLM's pretraining, which makes a truncation of memory necessary. Thus, it can lead to information loss due to the incompleteness of agent memory.
	Last but not least, it can lead to biases and unrobustness in LLM's inference.
	Specifically, a previous research~\cite{lost_in_the_middle} has shown that, the positions of text segments in a long context can greatly affect their utilization, so the memory in the long-context prompt can not be treated equally and stably. 
	All these drawbacks show the need to design extra memory modules for LLM-based agents, rather than straightforwardly concatenating all the information into a prompt.
	
	\textbf{Recent Interactions}.
	This method stores and maintains the most recently acquired memories using natural languages, thereby enhancing the efficiency of memory information utilization according to the Principle of Locality~\cite{principle_of_locality}.
	In task (B) of the example in \textbf{Section~\ref{basic}}, we can just remember Alice's preferences in the recent three years, and truncate the distant part, where the recent three years can be considered as the memory window size.
	
	In previous studies, there are various strategies to store recent textual memories.
	For example, SCM~\cite{scm} proposes a flash memory based on the cache mechanism, which preserves observations from the recent $t-1$ time steps, aimed at enhancing the recency of information.
	MemGPT~\cite{memgpt} considers the agent as an operating system, which can dynamically interact with users through a natural interface. It designs the working context to hold recent histories, as a part of virtual context management.
	In RecAgent~\cite{recagent}, the agents are designed to simulate user behaviors in movie recommendations. It stores some temporal information in short-term memory as an intermediate cache, which can simulate the memory mechanism of the human brain~\cite{ebbinghaus1885memory,murre2015replication}.
	These representative methods can dynamically update memories based on recent interactions, and pay more attention to the recent context that is important for the current stage.
	
	Caching the memory according to recency is an effective way to enhance memory efficiency, and it enables agents to focus more on the recent information. However, in long-term tasks, this method fails to access key information from distant memories. It can result in the loss of potentially crucial information that is not within the immediate cache window. In other words, emphasizing on recency can inherently neglect earlier, yet critical information, thus posing challenges in scenarios requiring a comprehensive understanding of past events.
	
	\textbf{Retrieved Interactions}.
	Unlike the above method which truncates memories based on time, this method typically selects memory contents based on their relevance, importance, and topics.
	It ensures the inclusion of distant but crucial memories in the decision-making process, thereby addressing the limitation of only memorizing recent information.
	In task (A) of the example in \textbf{Section~\ref{basic}}, Alice's preferences have been stored in the memory before this task. At [Step 2], the agent will retrieve the most relevant aspects of Alice's preferences from memory based on the query keyword "travel", obtaining Alice's scenic spot preference for ancient architectures. In general, retrieval methods will generate embeddings as indexes for memory entries during memory writing, along with recording auxiliary information to assist in retrieval. During memory reading, matching scores are calculated for each memory entry, and the top-$K$ entries will be used for the decision-making process of agents.
	
	In existing studies, most agents utilize retrieval methods to process the memory information.
	For example, \citet{generative_agents} first calculate the relevance between the current context and memory entries by cosine similarity, and obtain the importance and recency according to auxiliary information.
	MemoryBank~\cite{memorybank} employs a dual-tower dense retrieval model to find related information from past conversations. Each memory entry is encoded into an embedding and subsequently indexed by FAISS~\cite{faiss} to improve the efficiency of retrieval. When reading memories, the current context will be encoded as representations to obtain the most relevant memory.
	Moreover, RET-LLM~\cite{ret_llm} intends to design a write-read memory module for general usage. It utilizes Locality-Sensitive Hashing (LSH) to retrieve tuples with relative entries in the database to provide more information.
	In addition, ChatDB~\cite{chatdb} designs to utilize symbolic memory, and proposes to generate SQL statements to retrieve from database to obtain stored information. 
	
	The retrieval methods considerably depend on the accuracy and efficiency of obtaining expected information. An inaccurate retrieval strategy can potentially acquire unrelated information that is unhelpful for agent inference. And a heavy retrieval system can lead to large computational costs and long time latency, especially when handling massive information.
	Moreover, retrieval methods typically store homogeneous information inside the environment, where all the information is in a consistent form. For heterogeneous information outside the environment, it's difficult to directly apply the same method for memory storage. 
	
	\textbf{External Knowledge}.
	To obtain more information, some agents acquire external knowledge by invoking tools, with the aim of transforming additional relevant knowledge into their own memories for decision-making. 
	For instance, accessing external knowledge through Application Programming Interface (API) is a common practice~\cite{react,reflexion}. Nowadays, abundant public information, such as Wikipedia and OpenWeatherMap\footnote{\url{https://openweathermap.org}}, are available online (either free of charge or on a paying basis), and can be conveniently accessed through API calls.
	For instance, in [Step 2] of task (A) of the example in \textbf{Section~\ref{basic}}, external knowledge from the digital magazine is obtained with tool methods.
	
	In existing models, Toolformer~\cite{toolformer}
	proposes to teach LLM to use tools, which can acquire external knowledge for better solving tasks.
	Furthermore, ToolLLM~\cite{toolllm} empowers Llama~\cite{llama} with the ability to utilize more APIs in RapidAPI\footnote{\url{https://rapidapi.com/hub}} and to enable multi-tool usage, which provides a general interface to extend agents' ability.
	In TPTU~\cite{tptu}, the agents are incorporated in both task planning and tool usage, in order to tackle intricate problems. The follow-up work~\cite{tptu_v2} further improves its ability extensively like retrieval.
	In ToRA~\cite{tora}, the agents are required to solve mathematical problems. They utilize imitation learning to improve their ability to use program-based tools.
	
	The above methods significantly advance the capabilities of agents by allowing them to access external up-to-date and real-world information from diverse sources. However, the reliability of this information can be questionable due to potential inaccuracies and biases~\cite{survey_tool_learning}. Furthermore, the integration of tools into agents demands a comprehensive understanding to interpret the retrieved information across various contexts, which can incur higher computational costs and complications in aligning external data with internal decision-making processes. Additionally, utilizing external APIs brings forth concerns regarding privacy, data security, and compliance with usage policies, necessitating rigorous management and oversight~\cite{survey_tool_learning}.
	
	\subsubsection{Memory in Parametric Form}
	An alternative type of approaches is to represent memory in parametric form. They do not take up the extra length of context in prompts, so they are not constrained by the length limitations of LLM context. However, the parametric memory form is still under-researched, and we categorize previous works into two types: fine-tuning methods and memory editing methods.
	
	\textbf{Fine-tuning Methods.}
	Integrating external knowledge into the memory of agents is beneficial for enriching domain-specific knowledge on top of its general knowledge.
	To infuse the domain knowledge into LLMs, supervised fine-tuning is a common approach, which empowers agents with the memory of domain experts. It significantly improves the agent's ability to accomplish domain-specific tasks.
	In task (A) of the example in \textbf{Section~\ref{basic}}, the external knowledge of attractions from magazines can be fine-tuned into the parameters of LLMs prior to this task.
	
	In previous works, Character-LLM~\cite{character_llm} focuses on the role-play circumstance. It utilizes supervised fine-tuning strategies with role-related data (\textit{e.g.,} experiences), to endow agents with the specific traits and characteristics of the role.
	Huatuo~\cite{huatuo} intends to empower agents with professional ability in the biomedical domain. It tries to fine-tune Llama~\cite{llama} on Chinese medical knowledge bases.
	Besides, in order to create artificial doctors, DoctorGLM~\cite{doctorglm} fine-tunes ChatGLM~\cite{chatglm} with LoRA~\cite{lora}, and Radiology-GPT~\cite{radiology_gpt} improves domain knowledge on radiology analysis by supervised fine-tuning on an annotated radiology dataset. Moreover, InvestLM~\cite{investlm} collects investment data and fine-tunes it to improve domain-specific abilities on financial investment.
	
	The fine-tuning methods can effectively bridge the gap between general agents and specialized agents.
	It improves the capability of agents on the tasks that require high accuracy and reliability on domain-specific information.
	Nevertheless, fine-tuning LLMs for specific domains could potentially lead to overfitting, and it also raises concerns about catastrophic forgetting, where LLMs may forget the original knowledge because of updating their parameters.
	Another limitation of fine-tuning lies in the computational cost and time consumption, as well as the requirement of a large amount of data.
	Therefore, most fine-tuning approaches are applied to offline scenarios, and can seldom deal with online scenarios, such as fine-tuning with agent observations and trial experiences. Due to the frequent agent-environment interactions, it is unaffordable for the cost of backpropagation to fine-tune every step of the online and dynamic interactions.
	
	\textbf{Memory Editing Methods.}
	Apart from the fine-tuning approaches, another type of methods for infusing memory into model parameters is knowledge editing~\cite{knowledgeeditor,mend}.
	Unlike fine-tuning methods that extract patterns from certain datasets, knowledge editing methods specifically target and adjust only the facts that need to be changed. It ensures that unrelated knowledge remains unaffected. 
	Knowledge editing methods are more suitable for small-scale memory adjustments. Generally, they have lower computational costs, making them more suitable for online scenarios.
	In our example of task (B), Alice always watches movies at 9:00 PM from the agent's memory, but she may recently change her work and would not be empty at 9:00 PM. If so, the related memory (such as routines at 9:00 PM) should be edited, which can be implemented by knowledge editing methods.
	
	In previous studies, MAC~\cite{mac} intends to design an effective and efficient memory adaptation framework for online scenarios. It utilizes meta-learning to substitute the optimization step.
	PersonalityEdit~\cite{personalityedit} focuses on editing the personality of LLMs and agents, where it changes their traits based on theories such as the big-five factor.
	MEND~\cite{mend} utilizes the idea of meta-learning to train a lightweight model, which is capable of generating modifications for model parameters of a pre-trained language model.
	APP~\cite{app_paper} studies whether adding a new fact leads to catastrophic forgetting of existing facts. It focuses on the impact of neighbor perturbation on memory addition.
	Moreover, KnowledgeEditor~\cite{knowledgeeditor} trains a hyper-network to predict the modification of model parameters when injecting memory based on a learning-to-update problem formulation.
	\citet{dinm} propose a new optimization target to change the poisoning knowledge of LLM, and maintain the general performance at the same time. For LLM-based agents, the agents can change bad memory by knowledge editing, which can be considered as a type of forgetting mechanism. 
	
	Knowledge editing methods provide an innovative way to update the information stored within the parameters of LLMs. By specifically targeting and adjusting the facts, these methods can ensure the non-targeted knowledge unaffected during updates, thus mitigating the issue of catastrophic forgetting. Moreover, the targeted adjustment mechanism allows for more efficient and less resource-intensive updates, making knowledge editing an appealing choice for high-precision and real-time modifications. However, despite these promising developments, computational costs of meta-training and the preservation of unrelated memories remain significant challenges.
	
	\subsubsection{Advantages and Disadvantages of Textual and Parametric Memory}
	Textual memory and parametric memory have their strengths and weaknesses respectively, making them suitable for different memory contents and application scenarios. In this section, we discuss the advantages and disadvantages of these two forms of memory from various aspects.
	
	\textbf{Effectiveness.}
	The textual memory stores raw information about the agent-environment interactions, which is more comprehensive and detailed. However, it is constrained by the token limitation of LLM prompts, which makes the agent hard to store extensive information. 
	In contrast, the parametric memory is not limited by the prompt length, but it may suffer from information loss when transforming texts into parameters, and the complex memory training can bring additional challenges.
	
	\textbf{Efficiency.}
	For textual memory, each LLM inference requires to integrate memory into the context prompt, which leads to higher costs and longer processing times. In contrast, for parametric memory, the information can be integrated into the parameters of the LLM, eliminating the extra costs of these contexts. However, parametric memory takes additional costs in the writing process, but textual memory is easier to write, especially for small amounts of data. 
	In a nutshell, textual memory is more efficient in writing, while parametric memory is more efficient in reading.
	
	\textbf{Interpretability.}
	Textual memory is usually more explainable than the parametric one, since natural languages are the most natural and straightforward strategies for humans to understand, while parametric memory is commonly represented in latent space. 
	Nevertheless, such explainability is obtained at the cost of information density. This is because the sequences of words in textual memory are represented in a discrete space, which is not as dense as continuous space in parametric memory.
	
	In conclusion, the trade-offs between these two types of memories make them suitable for different applications. 
	For example, for the tasks that require recalling recent interactions, like conversational and context-specific tasks, textual memory seems more effective. For the tasks that require a large amount of memory, or well-established knowledge, parametric memory can be a better choice.
	
	\subsection{Memory Operations}
	We separate the entire procedure of memory into three operations: memory writing, memory management, and memory reading. These three typically collaborate to achieve memory function, providing information for LLM inference.
	We summarize previous works on memory operations in \textbf{Table~\ref{tab:memory_operation}}.
	
	\begin{table}[tbp]
		\centering
		\caption{
			Summarization of the memory operations.
			If a model does not have special designs on the memory operations, we use $\circ$ to label it, otherwise, it is denoted by $\checkmark$.
			\texttimes~means that the memory operations are not discussed in the paper.
		}
		\vspace{0.05cm}
		\resizebox{\linewidth}{!}{
			\begin{tabular}{c>{\centering\arraybackslash}p{2cm}>{\centering\arraybackslash}p{1.8cm}>{\centering\arraybackslash}p{1.8cm}>{\centering\arraybackslash}p{1.8cm}>{\centering\arraybackslash}p{2cm}}
				\hline
				\hline
				\multirow{2}[4]{*}{\textbf{Models}} & \multirow{2}[4]{*}{\textbf{Writing}} & \multicolumn{3}{c}{\textbf{Management}} & \multirow{2}[4]{*}{\textbf{Reading}} \bigstrut\\
				\cline{3-5}          &       & \textbf{Merging} & \textbf{Reflection} & \textbf{Forgetting} &  \bigstrut\\
				\hline
				MemoryBank~\cite{memorybank} & \checkmark & \checkmark & \checkmark & \checkmark & \checkmark \bigstrut[t]\\
				RET-LLM~\cite{ret_llm} & \checkmark & \texttimes & \texttimes & \texttimes & \checkmark \\
				ChatDB~\cite{chatdb} & \checkmark & \texttimes & \checkmark & \texttimes & \checkmark \\
				TiM~\cite{tim} & \checkmark & \checkmark & \texttimes & \checkmark & \checkmark \\
				SCM~\cite{scm} & \checkmark & \checkmark & \texttimes & \texttimes & \checkmark \\
				Voyager~\cite{voyager} & \checkmark & \texttimes & \checkmark & \texttimes & \checkmark \\
				MemGPT~\cite{memgpt} & \checkmark & \texttimes & \checkmark & \texttimes & \checkmark \\
				MemoChat~\cite{memochat} & \checkmark & \texttimes & \texttimes & \texttimes & \checkmark \\
				MPC~\cite{mpc} & \checkmark & \texttimes & \texttimes & \texttimes & \checkmark \\
				Generative Agents~\cite{generative_agents} & \checkmark & \texttimes & \checkmark & \checkmark & \checkmark \\
				RecMind~\cite{recmind} & $\circ$ & \texttimes & \texttimes & \texttimes & \checkmark \\
				Retroformer~\cite{retroformer} & \checkmark & \checkmark & \checkmark & \texttimes & $\circ$ \\
				ExpeL~\cite{expel} & \checkmark & \checkmark & \checkmark & \texttimes & $\circ$ \\
				Synapse~\cite{synapse} & \checkmark & \texttimes & \texttimes & \texttimes & \checkmark \\
				GITM~\cite{gitm} & $\circ$ & \checkmark & \checkmark & \texttimes & \checkmark \\
				ReAct~\cite{react} & $\circ$ & \texttimes & \texttimes & \texttimes & $\circ$ \\
				Reflexion~\cite{reflexion} & \checkmark & \checkmark & \checkmark & \texttimes & $\circ$ \\
				RecAgent~\cite{recagent} & \checkmark & \checkmark & \checkmark & \checkmark & \checkmark \\
				Character-LLM~\cite{character_llm} & \checkmark & \texttimes & \texttimes & \texttimes & $\circ$ \\
				MAC~\cite{mac} & \checkmark & \checkmark & \checkmark & \texttimes & \checkmark \\
				Huatuo~\cite{huatuo} & \checkmark & \texttimes & \texttimes & \texttimes & $\circ$ \\
				ChatDev~\cite{chatdev} & \checkmark & \texttimes & \checkmark & \texttimes & \checkmark \\
				InteRecAgent~\cite{interecagent} & \checkmark & \checkmark & \checkmark & \texttimes & \checkmark \\
				MetaAgents~\cite{metaagents} & \checkmark & \texttimes & \checkmark & \texttimes & \checkmark \\
				TPTU~\cite{tptu, tptu_v2} & $\circ$ & \texttimes & \checkmark & \texttimes & \checkmark \\
				MetaGPT~\cite{metagpt} & \checkmark & \texttimes & \checkmark & \texttimes & \checkmark \\
				S$^3$~\cite{s3} & \checkmark & \texttimes & \checkmark & \checkmark & \checkmark \\
				InvestLM~\cite{investlm} & \checkmark & \texttimes & \texttimes & \texttimes & $\circ$ \bigstrut[b]\\
				\hline
				\hline
			\end{tabular}%
		}
		\label{tab:memory_operation}%
		\vspace{-0.4cm}
	\end{table}%
	
	\subsubsection{Memory Writing}
	
	After the information is perceived by the agent, a part of it will be stored by the agent for further usage through the memory writing operation, and it is crucial to recognize which information is essential to store.
	Many studies choose to store the raw information, while others also put the summary of the raw information into the memory module.

	\textbf{Representative Studies.}
	In TiM~\cite{tim}, the raw information will be extracted as the relation between two entities, and stored in a structured database. When writing into the database, similar contents will be stored in the same group.
	In SCM~\cite{scm}, it designs a memory controller to decide when to execute the operations. The controller serves as a guide for the whole memory module.
	In MemGPT~\cite{memgpt}, the memory writing is entirely self-directed. The agents can autonomously update the memory based on the contexts.
	In MemoChat~\cite{memochat}, the agents summarize each conversation segment by abstracting the mainly discussed topics and storing them as keys for indexing memory pieces.

	\textbf{Discussion.}
	Previous research indicates that designing the strategy of information extraction during the memory writing operation is vital~\cite{memochat}. This is because the original information is commonly lengthy and noisy. Besides, different environments may provide various forms of feedback, and how to extract and represent the information as memory is also significant for memory writing.
	
	\subsubsection{Memory Management}
	For human beings, memory information is constantly processed and abstracted in the brains. The memory in the agent can also be managed by reflecting to generate higher-level memories, merging redundant memory entries, and forgetting unimportant, early memories.
	
	\textbf{Representative Studies.}
	In MemoryBank~\cite{memorybank}, the agents process and distill the conversations into a high-level summary of daily events, similar to how humans recall key aspects of their experiences.
	Through long-term interactions, they continually evaluate and refine their knowledge, generating daily insights into personality traits.
	In Voyager~\cite{voyager}, the agents are able to refine their memory based on the feedback of the environment.
	In Generative Agents~\cite{generative_agents}, the agents can reflect to get higher-level information, where the abstract thoughts are generated from agents. The reflection process will be activated when there are accumulated events that are enough to address.
	For GITM~\cite{gitm}, in order to establish common reference plans for various situations, key actions from multiple plans are further summarized in the memory module.
	
	\textbf{Discussion.}
	Most of the memory management operations are inspired by the working mechanism of human brains. With the strong capability of LLMs to simulate human minds, these operations can help the agents to better generate high-level information and interact with environments.

	\subsubsection{Memory Reading}
	When the agents require information for reasoning and decision-making, the memory reading operation will extract related information from memory for usage. Therefore, how to access the related information for the current state is important. Due to the massive quantity of memory entities, and the fact that not all of them are pertinent to the current state, careful design is required to extract useful information based on relevance and other task-orientated factors.
	
	\textbf{Representative Studies.}
	In ChatDB~\cite{chatdb}, the memory reading operation is executed by the SQL statements. These statements will be generated by agents as a series of Chain-of-Memory in advance.
	In MPC~\cite{mpc}, the agents can retrieve relevant memory from the memory pool. This method also proposes to provide Chain-of-Thought examples for ignoring certain memory.
	ExpeL~\cite{expel} utilizes the Faiss~\cite{faiss} vector store as the pool of memory, and obtains the top-$K$ successful trajectories that share the highest similarity scores with the current task.
	
	\textbf{Discussion.}
	To some extent, the memory reading and writing operations are collaborative, and the forms of memory writing greatly influence the methods of memory reading. For the forms of textual memory, most previous works use the text similarity and other auxiliary information for reading. For the forms of parametric memory, existing models may just utilize the updated parameters for inference, which can be seen as an implicit reading process.

	\section{How to Evaluate the Memory in LLM-based Agent}
	\label{sec:evaluation}
\begin{figure*}[tb]
		\centering
		\setlength{\fboxrule}{0.pt}
		\setlength{\fboxsep}{0.pt}
		\fbox{
			\includegraphics[width=\linewidth]{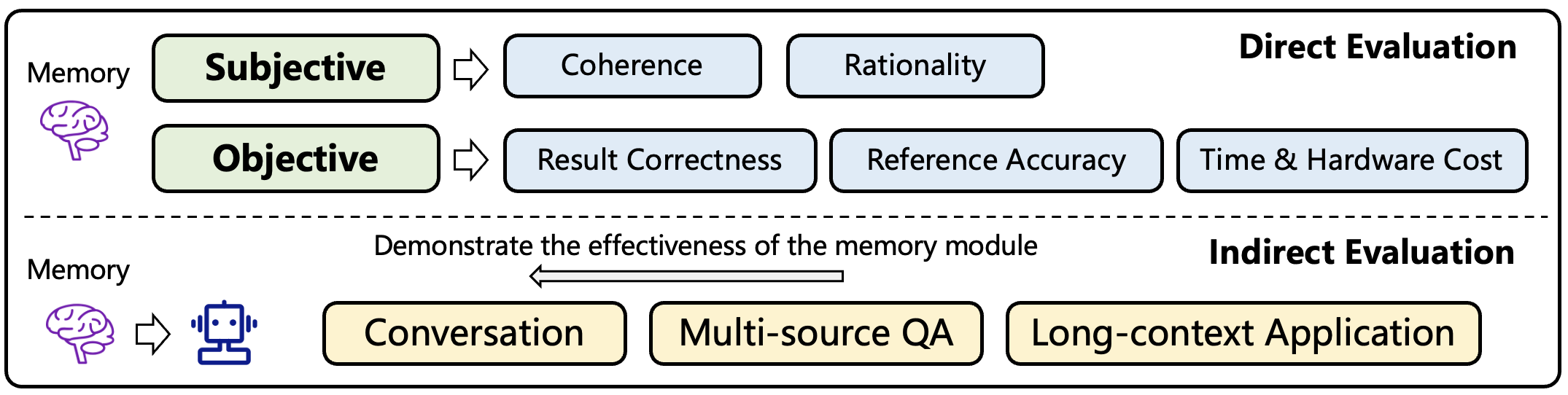}
		}
		\vspace{-0.4cm}
		\caption{An overview of the evaluation methods of the memory module.}
		\label{fig:evaluation}
		\vspace{-0.6cm}
	\end{figure*}

	How to effectively evaluate the memory module remains an open problem, where diverse evaluation strategies have been proposed in previous works according to different applications. To clearly show the common ideas of different evaluation methods, in this section, we summarize a general framework, which includes two broad evaluation strategies (see \textbf{Figure~\ref{fig:evaluation}} for an overview), that is, (1) direct evaluation, which independently measures the capability of the memory module. 
	(2) indirect evaluation, which evaluates the memory module via end-to-end agent tasks. 
	If the tasks can be effectively accomplished, the memory module is demonstrated to be useful. 
	
	\subsection{Direct Evaluation}
	This type of approaches regards the memory of the agents as a stand-alone component and evaluates its effectiveness independently.
	Previous studies can be categorized into two classes: subjective evaluation and objective evaluation. The subjective evaluation aims to measure memory effectiveness based on human judgments, which can be widely used in the scenarios that lack objective ground truths. Objective evaluation assesses memory effectiveness based on numerical metrics, which makes it easy to compare different memory modules.
	
	\subsubsection{Subjective Evaluation}
	In subjective evaluation, there are two key problems, that is, (1) what aspects should be evaluated and (2) how to conduct the evaluation process.
	To begin with, the following two aspects are the most common perspectives leveraged to evaluate the memory module.
	
	\textbf{Coherence.} This aspect refers to whether the recalled memory is natural and suitable for the current context. For example, if the agent is making a plan for Alice's travel, the memory should be related to her preference for traveling rather than working.
	In previous works, \citet{ret_llm} study whether the memory module could provide proper references among the ever-changing knowledge.
	\citet{scm} present some examples to demonstrate the relation between the current query and historical memory.
	\citet{memorybank} and \citet{tim} assess the coherence of responses that integrate context and retrieved memory by scoring labels. \citet{mpc} focus on the contradiction between the recalled memory and contexts.
	
	\textbf{Rationality.} This aspect aims to evaluate whether the recalled memory is reasonable. For example, if the agent is asked to answer "Where is the Summer Palace", the recalled memory should be "The Summer Palace is in Beijing" rather than "The Summer Palace is on the Moon". In previous works, \citet{mpc} ask crowd workers to directly score the rationality of the retrieved memory. \citet{memorybank} and \citet{tim} recruit human evaluators to check if the memory contains reasonable answers for the current question.
	
	As for how to conduct the evaluation process, there are two important problems.
	The first one is how to select the human evaluators. 
	In general, the evaluators should be familiar with the evaluation task, which ensures that the labeling results are convincing and reliable. 
	In addition, the backgrounds of the evaluators should be diverse to remove subjective biases of specific human groups. 
	The second problem is how to label the outputs of the memory module. Usually, one can either directly score the results~\cite{memorybank} or make comparisons between two candidates~\cite{recagent}.
	The former can obtain absolute and quantitative evaluation results, while the latter can remove the labeling noises when independently scoring each candidate. 
	In addition, the granularity of the ratings should also be carefully designed. Too coarse ratings may not effectively discriminate the capabilities of different memory modules, while too fine-grained ones may bring more effort for the workers to make judgments.
	
	In general, subjective evaluation can be used in a wide range of scenarios, where one just needs to define the evaluation aspects and let recruited workers make judgments. 
	This method is usually more explainable since the workers can provide the reasons for their judgments. 
	However, subjective evaluation is costly due to the need to employ human evaluators. Additionally, different groups of evaluators may have various biases, making the results difficult to reproduce and compare.
	
	\subsubsection{Objective Evaluation}
	In objective evaluation, previous work usually defines numeric metrics to evaluate the effectiveness and efficiency of the memory module.
	
	\textbf{Result Correctness.} This metric measures whether the agent can successfully answer pre-defined questions directly based on the memory module.
	For example, the question could be "Where did Alice go today?" with two choices "A: the Summer Palace" and "B: the Great Wall". Then, the agent should choose the correct answer based on the problem and its memory.
	The agent-generated answer will be compared with the ground truth. Formally, the accuracy can be calculated as
	$$
	\text{Correctness} = \frac{1}{N} \sum_{i=1}^N \mathbb{I}\left [a_i = \hat{a}_i \right ],
	$$
	where $N$ is the number of problems, $a_i$ represents the ground truth for the $i$-th problem, $\hat{a}_i$ means the answer given by the agent, and $\mathbb{I}\left [a_i = \hat{a}_i \right ]$ is the matching function commonly represented as
	$$
	\mathbb{I}\left [a_i = \hat{a}_i \right ] = \begin{cases}
		1 & \text{if } a_i = \hat{a}_i, \\
		0 & \text{if } a_i \neq \hat{a}_i.
	\end{cases}
	$$
	
	In previous works, \citet{chatdb} construct questions from past histories with annotated ground truths and calculate the accuracy of whether the recalled memory could match the correct answers. Similarly, \citet{memgpt} generate questions and answers that can only be derived from past sessions, and compare the responses from the agents with the ground truths to calculate the accuracy.
	
	\textbf{Reference Accuracy.} This metric evaluates whether the agent can discover relevant memory contents to answer the questions. 
	Different from the above metric, which focuses on the final results, reference accuracy cares more about the intermediate information to support the agent's final decisions. In specific, it compares the retrieved memory with the pre-prepared ground truth.
	For the above problem of "Where did Alice go today?", if the memory contents include (A) "Alice had lunch with friends at \textit{Wangfujing} today." and (B) "Alice had roast duck for lunch", then a better memory module should select (A) as a reference to answer the question.
	Usually, researchers leverage F1-score to evaluate the reference accuracy, which is calculated as
	$$
	\text{F1} = 2 \cdot \frac{\text{Precision} \cdot \text{Recall}}{\text{Precision} + \text{Recall}},
	$$
	where the precision and recall scores are calculated as
	$\text{Precision} = \frac{\text{TP}}{\text{TP}+\text{FP}}$ and $\text{Recall} = \frac{\text{TP}}{\text{TP}+\text{FN}}$. The \text{TP} represents the number of true positive memory contents, \text{FP} means the number of false positive memory contents, and \text{FN} indicates the number of false negative memory contents.
	In previous works, \citet{memochat} utilize F1-score to evaluate the retrieval process of the memory, and \citet{memorybank} focus on assessing whether related memory can be successfully retrieved.
	
	Result Correctness and Reference Accuracy are both utilized to evaluate the effectiveness of the memory module. Beyond effectiveness, efficiency is also an important aspect, especially for real-world applications.
	Therefore, we describe the evaluation of efficiency as follows.
	
	\textbf{Time \& Hardware Cost.} 
	The total time cost includes the time leveraged for memory adaption and inference.
	The adaptation time refers to the time of memory writing and memory management, while the inference time indicates the time latency of memory reading. In specific, the difference from the end time to the start time of memory operations can be considered as the time consumption.
	Formally, the average time consumption of each type of operation can be represented as
	$$
	\Delta \text{time} = \frac{1}{M} \sum_{i=1}^M t_i^{\text{end}} - t_i^{\text{start}},
	$$
	where $M$ represents the number of these operations, $t_i^{\text{end}}$ means the end time of the $i$-th operation, and $t_i^{\text{start}}$ indicates the start time of that operation. As for the computation overhead, it can be evaluated by the peak GPU memory allocation. In previous works, \citet{mac} utilize the peak memory allocation and adaptation time to assess the efficiency of memory operations. 
	
	Objective evaluation offers numeric strategies to compare different methods of memory, which is important to benchmark this field and promote future developments.

	\subsection{Indirect Evaluation}
	Besides the above method that directly evaluates the memory module, evaluating via task completion is also a popular evaluation strategy.
	The intuition behind this type of approaches is that if the agent can successfully complete a task that highly depends on memory, it suggests that the designed memory module is effective.
	In the following parts, we present several representative tasks that are leveraged to evaluate the memory module in indirect ways.

	\subsubsection{Conversation}
	Engaging in conversations with humans is one of the most important applications of agents, where memory plays a crucial role in this process. By storing context information in memory, the agents allow users to experience personalized conversations, thus improving users' satisfaction. Therefore, when other parts of the agents are determined, the performance of the conversation tasks can reflect the effectiveness of different memory modules.
	
	In the context of conversation, consistency and engagement are two commonly used methods to evaluate the effectiveness of the agents' memory.
	Consistency refers to how the response from agents is consistent with the context because dramatic changes should be avoided during the conversation. For example, \citet{memochat} evaluate the consistency of agents on interactive dialogues, using GPT-4 to score on the responses from agents.
	Engagement refers to how the user is engaged to continue the conversation. It reflects the quality and attraction of agents' responses, as well as the ability of agents to craft the personas for current conversations. For example, \citet{mpc} assess the engagingness of responses by SCE-p score, and \citet{memgpt} utilize CSIM score to evaluate the memory effect on increasing engagement of users.
	
	\subsubsection{Multi-source Question-answering}
	Multi-source questing-answering can comprehensively evaluate the memorized information from multiple sources, including inside-trial information, cross-trial information, and external knowledge. It focuses on the integration of memory utilization from various contents and sources.
	
	In previous works, \citet{react} evaluate the memory that integrates information from the task trial and the external knowledge from Wikipedia. Then, \citet{reflexion} and \citet{retroformer} further include the cross-trial information of the same task, where the memory is permitted to obtain more experiences from previous failed trials. Moreover, \citet{memgpt} allow agents to utilize the memory from multi-document information for question-answering.
	
	By evaluating multi-source question-answering tasks, the memory of agents can be examined on the capability of content integration from various sources. It also reveals the issue of the memory contradiction due to multiple information sources, and the problem of updated knowledge, which can potentially affect the performance of the memory module.
	
	\subsubsection{Long-context Applications}
	Beyond the above general applications, in many scenarios, LLM-based agents have to make decisions based on extremely long prompts. 
	In these scenarios, the long prompts are usually regarded as the memory contents, which play an important role in driving agent behaviors.
	
	In previous works, \citet{survey_long_context_llm} organize a comprehensive survey for long-context LLMs, which provides a summary of evaluation metrics on long-context scenarios. Moreover, \citet{zeroscrolls} propose a zero-shot benchmark for evaluating agents' understanding of long-context natural languages. 
	As for specific long-context tasks, long-context passage retrieval is one of the important tasks for evaluating the long-context ability of agents. It requires agents to find the correct paragraph in a long context that corresponds to the given questions or descriptions~\cite{longbench}. Long-context summarization is another representative task. It requests agents to formulate a global understanding of the whole context, and summarizes it according to the descriptions, where some metrics on matching scores like ROUGE can be utilized to compare the results with ground truths.
	
	The evaluation of long-context applications provides broader approaches to assess the function of memory in agents, focusing on practical downstream scenarios. The comprehensive benchmarks~\cite{zeroscrolls,longeval} also provide an objective assessment for the ability of long-context understanding.
	
	\subsubsection{Other Tasks}
	In addition to the above three types of major tasks for indirect evaluation, there are also some other metrics in general tasks that can reveal the effectiveness of the memory module.
	
	Success rate refers to the proportion of tasks that agents can successfully solve. For \citet{react,reflexion} and \citet{expel}, they assess how many spacial tasks can be correctly completed through reasoning and memory in AlfWorld~\cite{alfworld}. In \citet{gitm}, they evaluate the success rate of producing different items in Minecraft to show the effect of memory. Moreover, \citet{reflexion} measure the success rate of passed problems by generated codes, and \citet{synapse} calculate the success rate of computer control and accuracy of element selection to show the function of trajectory-as-exemplar memory.
	Exploration degree typically appears in exploratory games, which reflects the extent that agents can explore the environment. For example, \citet{voyager} compare the numbers of distinct items explored in Minecraft to reflect the skill learning in memory.
	
	In fact, nearly all the memory-equipped agents can evaluate the effect of memory by ablation studies, comparing the performance between with/without memory modules. The evaluation on specific scenarios can better reflect the significance of memory for the downstream applications practically.
	
	\subsection{Discussions}
	Compared with direct evaluation, indirect evaluation via specific tasks can be easier to conduct, since there are already many public benchmarks. However, the performance on tasks can be attributed to various factors, and memory is only one of them, which may make the evaluation results biased.
	By direct evaluation, the effectiveness of the memory module can be independently evaluated, which improves the reliability of the evaluation results. However, to our knowledge, there are no open-sourced benchmarks tailored for the memory modules in LLM-based agents.
	
	\section{Memory-enhanced Agent Applications}
	\label{sec:application}
	
	Recently, LLM-based agents have been investigated across a wide variety of scenarios, facilitating societal advancement. In general, most LLM-based agents are equipped with memory modules. However, the specific effects undertaken by these memory components, the particular information they store, and the implementation methods they use, vary across different applications. In order to provide insights for the design of memory functionalities in LLM-based agents, in this section, we review and summarize how memory mechanisms are manifested in LLM-based agents across various application scenarios. In specific, we categorize them into several classes: role-playing and social simulation, personal assistant, open-world games, code generation, recommendation, expert systems in specific domains, and other applications. The summarization is shown in \textbf{Table~\ref{tab:application}}.

	\begin{table}[tbp]
		\centering
		\caption{Summarization of memory-enhanced agents applications.}
		\resizebox{\linewidth}{!}{
			\begin{tabular}{c|>{\centering\arraybackslash}p{4cm}|c|>{\centering\arraybackslash}p{4cm}}
				\hline
				\hline
				\textbf{Applications} & \textbf{Models} & \textbf{Applications} & \textbf{Models} \bigstrut\\
				\hline
				\multirow{5}[2]{*}{Role-playing} & Character-LLM~\cite{character_llm} & \multirow{5}[2]{*}{Code Generation} & RTLFixer~\cite{rtlfixer} \bigstrut[t]\\
				&  ChatHaruhi~\cite{chatharuhi} &       & GameGPT~\cite{gamegpt} \\
				& RoleLLM ~\cite{rolellm} &       &   ChatDev~\cite{chatdev} \\
				&  NarrativePlay~\cite{narrativeplay} &       & MetaGPT~\cite{metaagents} \\
				&  CharacterGLM~\cite{characterglm} &       & CodeAgent~\cite{codeagent} \bigstrut[b]\\
				\hline
				\multirow{5}[4]{*}{Social Simulation} & Generative Agents~\cite{generative_agents} & \multirow{4}[2]{*}{Recommendation} & RecAgent~\cite{recagent} \bigstrut[t]\\
				& Lyfe Agents~\cite{lyfe_agents} &       & InteRecAgent~\cite{interecagent} \\
				& S$^3$~\cite{s3} &       & RecMind~\cite{recmind} \\
				&  MetaAgents~\cite{metaagents} &       & AgentCF~\cite{zhang2023agentcf} \bigstrut[b]\\
				\cline{3-4}      &  WarAgent~\cite{waragent} & \multirow{6}[4]{*}{Medicine} & Huatuo~\cite{huatuo} \bigstrut\\
				\cline{1-2}\multirow{9}[4]{*}{Personal Assistant} & MemoryBank~\cite{memorybank} &       & DoctorGLM~\cite{doctorglm} \bigstrut[t]\\
				& RET-LLM~\cite{ret_llm} &       & Radiology-GPT~\cite{radiology_gpt} \\
				& MemoChat~\cite{memochat} &       & \citet{cmedknowqa} \\
				& MemGPT~\cite{memgpt} &       & EHRAgent~\cite{ehragent} \\
				& MPC~\cite{mpc} &       & ChatDoctor~\cite{chatdoctor} \bigstrut[b]\\
				\cline{3-4}      & AutoGen~\cite{autogen} & \multirow{5}[4]{*}{Finance} & InvestLM~\cite{investlm} \bigstrut[t]\\
				& ChatDB~\cite{chatdb} &       & TradingGPT~\cite{tradinggpt} \\
				& TiM~\cite{tim} &       & QuantAgent~\cite{quantagent} \\
				& SCM~\cite{scm} &       &  FinMem~\cite{finmem} \bigstrut[b]\\
				\cline{1-2}\multirow{4}[4]{*}{Game} & Voyager~\cite{voyager} &       & \citet{koa2024learning} \bigstrut\\
				\cline{3-4}      & GITM~\cite{gitm} & \multirow{3}[2]{*}{Science} & Chemist-X~\cite{chemistx} \bigstrut[t]\\
				& JARVIS~\cite{jarvis1} &       & ChemDFM~\cite{chemdfm} \\
				& LARP~\cite{larp} &       &  MatChat~\cite{matchat} \bigstrut[b]\\
				\hline
				\hline
			\end{tabular}
		}	
		\label{tab:application}
		\vspace{-0.6cm}
	\end{table}%

	\subsection{Role-playing and Social Simulation}
	
	Role-playing represents a classic application of LLM-based agents, where memory plays a crucial role inside the agents. It endows roles with distinct characteristics, differentiating them from one another. Many previous studies have explored methods for constructing role memories~\cite{character_llm,chatharuhi,rolellm,narrativeplay,characterglm}. \citet{character_llm} construct the memory of roles by experience uploading, which utilizes SFT to inject memory into model parameters. \citet{chatharuhi} enhance large language models for role-playing via an improved prompt and the character memory extracted from scripts, where user queries and chatbot’s responses are concatenated to form a sequence as memory. \citet{rolellm} infuse role-specific knowledge and episode memories into LLM-based agents, where context QA pairs are concatenated to form episode memory. \citet{narrativeplay} aim to generate human-like responses, guided by personality traits extracted from narratives, which can be stored and retrieved by relevance and importance. \citet{characterglm} generate
	character-based dialogues for different roles and empower LLM-based agents with corresponding styles by SFT.
	
	Social simulation is basically an extension of role-playing, which focuses more on multi-agent modeling. The memory module is an important component for such applications, which helps to accurately simulate human dynamic behaviors. 
	In previous studies, \citet{lyfe_agents} propose a Summarize-and-Forget memory mechanism for better self-monitoring in social scenarios. \citet{s3} focus on social network simulation systems. Each agent in the system has a memory pool, which consists of diverse user messages from online platforms to identify the user. \citet{li2023large} maintain conversation contexts, encompassing the economic environment and agent decisions from previous months, in order to simulate the impact of broad macroeconomic trends on agents’ decision-making and to make the agents grasp market dynamics. \citet{metaagents} simulate the job-seeking scenario in human society, where the memory of agents includes profiles and goals initially and is further enriched with other information, like dialogues and personal reflections. \citet{waragent} simulate the decisions and consequences of the participating countries in the wars, where the conversations of agents are continuously maintained into memory.
	
	There are several insights in designing an agent's memory for role-play and social simulation. First, the memory should be consistent with the roles' characteristics, which can be used to identify each role and distinguish it from the others. This is crucial for improving the realism of role-play and the diversity of social simulation. Second, the memory should appropriately influence the subsequent actions of the agent to ensure the consistency and rationality of its behaviors.
	Additionally, for humanoid agents, their memory mechanisms should align with the features of human memory, such as forgetting and long/short-term memory, which should refer to the theories of cognitive psychology.
	
	\subsection{Personal Assistant}
	
	LLM-based agents are well-suited for creating personal assistants, such as agents capable of engaging in long-term conversations with users~\cite{memochat,mpc,autogen}, as well as those tasked with automatically seeking information~\cite{kwaiagents}.
	These agents often need to memorize previous dialogues to maintain the consistency, and remember critical styles and events to generate more personalized and relevant responses.
	\citet{memochat} maintain the context consistency for dialogues by saving contents and information of conversations, which helps to find proper relevant information by retrieval. \citet{mpc} summarize conversations to extract important information, store it, and retrieve it for future inference. \citet{kwaiagents} focus on information-seeking tasks, which design memory modules to store user's context information, and empower external knowledge with tool usage. \citet{autogen} retain important context as memory to maintain conversation consistency. 
	
	In summary, most memory implementations for personal assistants adopt retrieval methods in textual form, because they are better at finding relevant information from pieces of conversations.
	For the memory storage, the agent should remember the factual information during user-agent interactions, as well as the personal style of users, in order to generate responses that are tailored to the user's situation.
	Additionally, when recalling memories, the agent should identify and retrieve the memory that is relevant to the current query and context. This principle can enable the agent to correctly understand the user's requirement, and maintain the consistency in conversations.

	\subsection{Open-world Game}
	For games and open-world exploration, LLM-based agents always maintain post observations as task contexts, and store experiences in previous successful trials.
	By leveraging past experiences, agents can avoid making the same mistakes repeatedly and achieve a high-level understanding of environments, thus exploring more effectively.
	Some of them can acquire external databases or APIs to obtain general knowledge~\cite{voyager,gitm,jarvis1,larp}.
	\citet{voyager} save obtained skills into memory for further usage in Minecraft. \citet{gitm} store and retrieve successful trajectories as examples for similar tasks, and utilize external Minecraft Wiki by API calls. \citet{jarvis1} construct multimodal memory as a knowledge library and provide examples for prompt by retrieving interactive experiences. \citet{larp} maintain working memory for decision-making, save and retrieve relevant past experiences, and implement external datasets for general knowledge.
	
	In summary, no matter inside-trial or cross-trial information, the key aspect of memory is to reflect on past interactions and draw experiences that can be applied to the subsequent exploration. In addition to accumulating experience through self-involving trials, absorbing external knowledge as part of the agent's memory is also an important way to enhance the exploratory capabilities of the agent.

	\subsection{Code Generation}
	In the scenario of code generation, LLM-based agents can search relevant information from the memory, thereby obtaining more knowledge for development. They can save previous experiences for future problems, and also maintain context in conversational development interfaces~\cite{rtlfixer,gamegpt,chatdev,metaagents}.
	\citet{rtlfixer} construct an external non-parametric memory database, which stores the compiler errors and human expert instructions for automatic syntax error fixing. In \cite{gamegpt}, personal information will be stored in the memory, and helps in retaining context and knowledge for decision-making. \citet{chatdev} adopt multi-agents to develop software, where each role maintains a memory to store the past conversations with other roles. \citet{metaagents} also focus on software development, and the agent can retrieve its historical records preserved in memory when errors occur. \citet{codeagent} can search relevant information when they face problems on code generation. 
	
	By leveraging external resources, the agents can learn from code-related knowledge and store it into their memory, thereby enhancing the capabilities of code generation. In addition, the memory can improve the continuity and consistency in code generation. By integrating contextual memory, the agent can better understand the requirements for software development, thereby enhancing the coherence of the generated code. Furthermore, the memory is also crucial for the iterative optimization of code, as it can identify the developer's targets based on the histories.
	
	\subsection{Recommendation}
	
	In the field of recommendation, some previous works focus on simulating users in recommender systems~\cite{recagent,interecagent}, where the memory can represent the user profiles and histories in the real world.
	Others try to improve the performance of recommendation, or provide other formats of recommendation interfaces~\cite{zhang2023agentcf,recmind}.
	\citet{recagent} simulate user behaviors in recommendation scenarios to generate data for recommender systems, and the agents store past observations and insights into a hierarchical memory. In \citet{interecagent}, the memory in LLM-based agents can archive the user's conversational history over extended periods, as well as capture the most recent dialogues pertinent to the current prompt, to simulate interactive recommender systems. It also uses an actor-critic reflection to improve the robustness of agents. 
	Item agents and user agents are equipped with different memories in \cite{zhang2023agentcf}, where item agents are endowed with dynamic memory modules designed to capture and preserve information pertinent to their intrinsic attributes and the inclinations of their adopters. For user agents, the adaptive memory updating mechanism plays a pivotal role in aligning the agents' operations with user behaviors and preferences.
	\citet{recmind} memorize individualized user information like reviews or ratings for items, and acquire domain-specific knowledge and real-time information by web searching tools.
	
	For both simulating users in recommender systems and capturing their preferences, retaining personalized information through memory is essential. A critical challenge lies in how to align the personalized information and feedback with LLMs, and store them into the memory of agents. It is also an important task for bridging the gap between conventional recommendation models and LLMs.
	
	\subsection{Expert System in Specific Domains}
	
	\textbf{Medicine Domain.}
	In the field of medicine, most of the previous works empower LLM-based agents with external knowledge in their memory~\cite{huatuo,doctorglm,radiology_gpt,cmedknowqa,chatdoctor}. \citet{huatuo} fine-tune LLaMA~\cite{llama} with medical knowledge graph CMeKG~\cite{cmekg} in QA form, in order to enhance their medical domain knowledge. \citet{doctorglm} adopt LoRA~\cite{lora} to efficiently fine-tune on foundation models for healthcare. \citet{cmedknowqa} empower LLM-based agents to acquire text-based external knowledge as reasoning reference. Besides, \citet{ehragent} build memory upon the most relevant successful cases from past experiences, and use similarity metric for the retrieval of relevant questions in the medicine domain.
	
	\textbf{Finance Domain.}
	Some previous works also apply LLM-based agents in the finance domain, whose memory can store financial knowledge~\cite{investlm}, market information~\cite{tradinggpt,finmem}, and successful experiences~\cite{koa2024learning,quantagent}. \citet{investlm} construct financial investment dataset to fine-tune LLaMA~\cite{llama} to empower knowledge on investment. \citet{tradinggpt} design a layered-memory structure to store different types of marketing information. \citet{quantagent} record the ongoing interaction like exchanges and information to ensure consistent response, and record prior outputs as experiences for retrieving relevant examples to provide a diverse learning context for agents. \citet{koa2024learning} store past price movement and explanations, and generate reflections on previous trials. \citet{finmem} adopt a layered memory mechanism to provide abundant information for reasoning.
	
	\textbf{Science.}
	In the domain of science, some existing works design LLM-based agents with a large amount of knowledge in memory to solve problems~\cite{chemistx,chemdfm,matchat}. \citet{chemistx} include molecule database and online literature as external knowledge for memory in LLM-based agents, and retrieve them when they need related information. \citet{chemdfm} and \citet{matchat} empower domain knowledge by fine-tuning in Chemistry and structured materials respectively.
	
	To build an expert system based on agents in a specific vertical domain, it is necessary to retain the domain-specific knowledge in their memory. However, there are several challenges. First, domain knowledge is specialized and requires higher accuracy, leading to difficulties in constructing memory storage. Second, domain knowledge is often time-sensitive, which can become outdated in the future.
	Therefore, the memory needs to be partially updated when some of the knowledge has been out-of-date.
	Furthermore, the substantial volume of domain knowledge makes it difficult to recall from memory based on the current query.
	
	\subsection{Other Applications}
	There are some other applications of memory in LLM-based agents. \citet{rcagent} focus on the task of cloud root cause analysis, using memory to store framework rules, task requirements, tools documentation, few-shot examples, and agent observations. \citet{agent_om} solve the problem of ontology matching. The agents save conversational dialogues and construct a rational database for retrieving external knowledge. \citet{dilu} investigate autonomous driving, whose memory module is constructed by a vector database and contains the experiences from past driving scenarios. \citet{xuat} propose to improve user acceptance testing, which employs a self-reflection mechanism. After each trial, the operation agent summarizes the conversation and updates the memory pool, until the goal of the current step is accomplished. 
	
	For different applications, the focus of memory varies, as it inherently serves the downstream tasks. Therefore, the design should also consider the requirements of tasks.

	\section{Limitations \& Future Directions}
	\label{sec:future}
	
	\subsection{More Advances in Parametric Memory}
	
	At present, the memory of LLM-based agents is predominantly in textual form, especially for contextual knowledge such as observation records, trial experiences, and textual knowledge databases. Although textual memory possesses the advantages of being interpretable and easy to expand and edit, it also implies a sacrifice in efficiency compared to parametric memory. Essentially, parametric memory boasts a higher information density, expressing semantics through continuous real-number vectors in a latent space, whereas textual memory employs a combination of tokens in a discrete space for semantic expression. Thus, parametric memory offers a richer expressive space, and its soft encoding is more robust compared to the hard-coded form of token sequences.
	Additionally, parametric memory is more storage-efficient, where it does not require the explicit storage of extensive texts, similar to a knowledge compression process. As for the memory management, such as merging and reflection, parametric memory does not necessarily design manual rules like textual memory does, but can employ optimization methods to learn these processes implicitly. Moreover, pluggable parametric memory is similar to a digital life card, capable of endowing agents with the requisite characteristics.
	For example, Huatuo~\cite{huatuo} aims to enhance agents with expertise in the biomedical field by refining the Llama~\cite{llama} model on Chinese medical knowledge bases. MAC~\cite{mac} is designed to create a parametric memory adaptation framework suitable for online settings, employing meta-learning techniques to replace the traditional optimization phase.

	Although parametric memory holds great prospects, it currently faces numerous challenges. Foremost among these is the issue of efficiency: how to effectively transform textual information into parameters or modifications of parameters is a critical question. Presently, researchers can transfer vast amounts of domain knowledge into the parameters of LLMs by SFT. However, it is time-consuming and requires extensive text corpus, making it unsuitable for situational knowledge. One viable approach is to employ meta-learning to let models learn to memorize.
	For example, MEND~\cite{mend} leverages the method of meta-learning to train a compact model that has the ability to produce adjustments for the parameters of a pre-trained language model.
	Moreover, the lack of interpretability associated with parametric memory can be a hindrance, especially in domains requiring high levels of trust, such as medicine. Therefore, enhancing the credibility and interpretability of parametric memory is an urgent issue that needs to be addressed.

	\subsection{Memory in LLM-based Multi-agent Applications}
	
	The exploration of memory mechanisms within LLMs has burgeoned into the dynamic domain of multi-agent systems (MAS), marking significant advancements in the realms of synchronization, communication, and the management of information asymmetry.
	One pivotal aspect that emerges in the cooperative scenarios is memory synchronization among agents. This process is fundamental for establishing a unified knowledge base, ensuring consistency in decision-making across different agents. For example, \citet{chen2023scalable} emphasize the significance of integrating synchronized memory modules for multi-robot collaboration.
	Another important aspect is the communication among agents, which heavily relies on memory for maintaining context and interpreting messages. For example, \citet{mandi2023roco} illustrate memory-driven communication frameworks that foster a common understanding among agents.
	In addition to cooperative scenarios, some studies also focus on competitive scenarios, and the information asymmetry becomes a crucial issue~\cite{light2023text}.
	
	Looking ahead, the advancement of memory in LLM-based MAS is poised at the confluence of technological innovation and strategic application. It beckons the exploration of novel memory modules that can further enhance agent synchronization, enable more effective communication, and provide strategic advantages in information-rich environments. The development of such memory models would not only necessitate addressing the current challenges of memory integration and management, but also explore the untapped potentials of memory in facilitating more robust, intelligent, and adaptable MAS. As evidenced by pioneering research, the evolving landscape of LLM-based MAS sets a promising stage for future innovations in memory utilization and management. This exploration is expected to unravel new dimensions of memory integration, pushing the boundaries of what is currently achievable and setting new benchmarks in the realm of MAS.
	
	\subsection{Memory-based Lifelong Learning}
	Lifelong learning is an advanced topic in artificial intelligence, extending the learning capabilities of agents across their life-long span~\cite{survey_life_long_learning}. Agents can continuously interact with their environment, persistently observe environments, and acquire external knowledge, enabling a mode of enhancement like humans. The memory of an agent is key to achieving lifelong learning, as it needs to learn to store and apply the past observations. Lifelong learning in LLM-based agents holds significant practical value, such as in long-term social simulations and personal assistance. However, it also faces several challenges. Firstly, lifelong learning is temporal, necessitating that an agent's memory captures temporality. This temporality could cause interactions between memories, such as memory overlap. Furthermore, due to the extended period of lifelong learning, it needs to store a vast amount of memories and retrieve them when needed, possibly incorporating a certain mechanism for forgetting.
	
	\subsection{Memory in Humanoid Agent}
	
	A humanoid agent refers to an agent designed to exhibit behaviors consistent with humans, thereby facilitating applications in social simulation, studies of human behavior, and role-playing. Unlike task-oriented agents where greater capability is typically preferred, the proficiency of a humanoid agent should closely mimic that of humans. Consequently, the memory of humanoid agents should align with human cognitive processes, adhering to psychological principles such as memory distortion and forgetfulness. Additionally, humanoid agents should possess knowledge boundaries, meaning that their knowledge should correspond to that of the entity they replicate. For instance, in role-playing scenarios, an agent embodying a child should not possess an understanding of advanced mathematical concepts or other complex knowledge beyond what is typical for that age~\cite{aher2023using}.
	
	\section{Conclusion}
	\label{sec:conclusion}
	
	In this survey, we provide a systematical review on the memory mechanism of LLM-based agents, where we focus on three key problems including "What is", "Why do we need" and "How to design and evaluate" the memory module in LLM-based agents. 
	To show the importance of the agent's memory, we also present many typical applications, where the memory module plays an important role.
	We believe this survey can offer valuable references for newcomers to this domain, and also hope it can inspire more advanced memory mechanisms to enhance LLM-based agents.

	\section*{Acknowledgement}
	We thank Lei Wang for his proofreading and valuable suggestions to this survey.

	\bibliographystyle{unsrtnat}
	\bibliography{paper}

\end{document}